\documentclass{article}

\usepackage[main, final]{neurips_2026}

\usepackage[utf8]{inputenc} 
\usepackage[T1]{fontenc}    
\usepackage{hyperref}       
\usepackage{url}            
\usepackage{booktabs}       
\usepackage{amsfonts}       
\usepackage{nicefrac}       
\usepackage{microtype}      
\usepackage{xcolor}         

\usepackage{algorithm}
\usepackage{algorithmic}
\usepackage{multirow} 
\usepackage{lscape} 
\usepackage{pdfpages} 
\usepackage{longtable} 
\usepackage[T1]{fontenc} 
\usepackage{graphicx} 
\usepackage{caption}
\usepackage{subcaption}
\usepackage[most]{tcolorbox} 
\usepackage{CJKutf8}
\usepackage{amsmath}
\usepackage{amssymb}
\usepackage{mathtools}
\usepackage{amsthm}
\makeatletter

\newcommand{\Rmnum}[1]{\expandafter\@slowromancap\romannumeral #1@}
\makeatother
\usepackage{enumitem}
\usepackage[utf8]{inputenc} 
\usepackage[T1]{fontenc}    
\usepackage{url}            
\usepackage{amsfonts}       
\usepackage{nicefrac}       
\usepackage{microtype}      
\usepackage{xcolor}         
\usepackage{colortbl}
\usepackage{threeparttable}
\usepackage{fixltx2e}
\definecolor{MordantRed}{HTML}{D74D4D}
\definecolor{MordantGreen}{HTML}{68BB7E}
\usepackage[table]{xcolor}

\definecolor{codegreen}{rgb}{0,0.6,0}
\definecolor{codegray}{rgb}{0.5,0.5,0.5}
\definecolor{codepurple}{rgb}{0.58,0,0.82}
\definecolor{backcolour}{rgb}{0.95,0.95,0.92}
\definecolor{prollm}{RGB}{242, 242, 242} 
\definecolor{lrsa}{RGB}{242, 242,254} 
\definecolor{osgm}{RGB}{255, 253, 232} 
\definecolor{rag}{RGB}{245, 255, 243} 
\definecolor{KDASO}{RGB}{242, 176, 166}

\theoremstyle{plain}

\theoremstyle{definition}

\theoremstyle{remark}

\usepackage[toc,page]{appendix} 
\usepackage{tocloft} 
\usepackage{titletoc}
\title{KDASO: Knowledge-Data Dual-Driven Automated Skill Optimization for Liability Adjudication Tasks}

\author{%
\textbf{Zi-Jian Cheng\textsuperscript{1,2}\footnotemark[1] },
\textbf{Weiming Wu\textsuperscript{2}\footnotemark[1] },
\textbf{Jie Meng\textsuperscript{3}},
\textbf{Zhen Peng\textsuperscript{3}},
\textbf{Shan Huang\textsuperscript{3}},
\\
\textbf{Qun Li\textsuperscript{3}},
\textbf{Guobing Wu\textsuperscript{3}},
\textbf{Lan-Zhe Guo\textsuperscript{1,2}\footnotemark[2]}
\\
 \textsuperscript{1}National Key Laboratory for Novel Software Technology, Nanjing University
 \\
 \textsuperscript{2}School of Intelligence Science and Technology, Nanjing University
 \\
 \textsuperscript{3}DiDichuxing Co. Ltd, Beijing, China
\\
\texttt{wuwm23@smail.nju.edu.cn} \\
\texttt{\{chengzj,guolz\}@lamda.nju.edu.cn}\\
\texttt{\{jmengjie,zhenpeng,lattehuang,liquntracy,wuguobin\}@didiglobal.com}\\
}

\begin{document}

\maketitle
\renewcommand{\thefootnote}{\fnsymbol{footnote}}
\footnotetext[1]{ Equal contribution.}
\footnotetext[2]{ Corresponding author.}
\begin{abstract}
Liability adjudication tasks (e.g., legal charge determination) require analyzing event descriptions against domain-specific skills to determine whether a party should bear liability. While Large Language Models (LLMs) have shown promise in this area by leveraging executable skills for interpretable reasoning, existing approaches face three critical limitations: \textbf{skill libraries become stale as real-world scenarios dynamically evolve}, \textbf{skill descriptions often suffer from a semantic execution gap where LLMs deviate from intended expert logic}, and \textbf{growing skill collections exceed context budgets causing critical skill omission}. To address these challenges, we propose \textbf{K}nowledge-\textbf{D}ata Dual-Driven \textbf{A}utomated \textbf{S}kill \textbf{O}ptimization (KDASO), which continuously evolves a skill library through neighborhood-conditioned induction from recent adjudication data, execution-aligned refinement based on LLM trajectory analysis, and context-budgeted routing for scalable skill selection. Experiments across ride-hailing adjudication, legal liability determination, and logical reasoning judgment demonstrate that KDASO significantly outperforms current general-purpose and domain-specific baselines. KDASO further achieves 8.96\% improvement when transferring skills to concept-similar scenarios and maintains consistent performance gains through continuous optimization, underscoring the high transferability and stability of generated skills.
\end{abstract}

\section{Introduction}
\label{sec:introduction}

Liability adjudication~\cite{endicott2007adjudication, wu2026progressive}, as a prevalent and critical task in the real world, refers to the analysis of event descriptions based on domain-specific skills to determine whether a particular party should bear liability~\cite{hettiarachchi2025code}. Such tasks exist in various real-world scenarios, including ride-hailing dispute resolution and legal liability determination. For example, in a dispute over property management fees, the property owner is adjudicated not liable for breaching the payment obligation after executing a service-verification skill demonstrating that the property management company failed to provide the agreed-upon services as stipulated in the Civil Code.

Traditional algorithms for liability adjudication have primarily relied on logic-reasoning approaches~\cite{liu2025logical, yang2025neuro} and semantic-analysis methods~\cite{patil2023survey, aftan2023survey}. Logic-reasoning approaches rely on logical structures of skills to yield deductions~\cite{andreka2017new}, while semantic-analysis methods predict liability through similarity computation~\cite{chandrapati2024descriptive} or classifiers~\cite{liu2025simple}. However, these algorithms can only make adjudications based on fixed rules rather than optimizing the skills themselves, and they cannot generate interpretable reasoning trajectories. Recently, LLMs have emerged as a promising solution, offering a new paradigm for liability adjudication. Trained on vast domain-specific corpora, LLMs possess extensive embedded adjudication knowledge, which has sparked significant interest in leveraging them for this task~\cite{zhou2025lawgpt, yang2026appellategen}, particularly through Retrieval-Augmented Generation (RAG)~\cite{lewis2020retrieval, zhao2026retrieval}. 

These LLM-based approaches enhance the adjudication process by incorporating domain-specific priors into the reasoning pipeline, guiding evidence analysis with divergent skills that reflect varied expert perspectives, and providing interpretable reasoning trajectories that make the decision process transparent and traceable. Despite this promise, we identify three fundamental challenges that limit the effectiveness of skill-based LLM methods in liability adjudication. These challenges stem not from engineering imperfections, but from core tensions in how skills are acquired, executed, and deployed at scale, tensions that are intrinsic to the skill-based paradigm itself.

\textbf{Challenge 1: Skill Coverage under Distribution Shift.} Real-world liability scenarios are dynamically evolving. Emerging fraud behaviors, regulatory changes, and novel dispute patterns continually introduce decision boundaries that existing skill libraries fail to cover~\cite{cheng2025fully}. This creates a \emph{skill coverage gap} between the static knowledge embedded in the library and the shifting data distribution of incoming cases. A natural response is to induce new skills from recent adjudication data, yet this raises a fundamental learning dilemma: inducing a skill from a single case risks capturing idiosyncratic patterns that lack statistical validity, while inducing from the full dataset is intractable due to case heterogeneity and the excessive context length required to process all instances simultaneously.

\textbf{Challenge 2: Skill Execution Alignment.} Even when a skill is logically comprehensible to human experts, LLMs may interpret its natural language formulation inconsistently due to divergent priors, ambiguous operator precedence, or implicit assumptions. This creates a \emph{semantic execution gap} where the LLM's actual reasoning trajectory deviates from the expert's intended logic, even if the final label happens to be correct. Current approaches treat skills as static artifacts and do not account for the fact that LLM execution is inherently stochastic and sensitive to phrasing. Without a feedback mechanism that observes the LLM's execution trajectories and adjusts the skill formulation accordingly, the system cannot guarantee stable or faithful skill execution.

\textbf{Challenge 3: Skill Scalability under Context Budget.} As the skill library grows to cover diverse scenarios, feeding the entire collection into LLM alongside a test case becomes infeasible due to finite context windows. This inevitably triggers skill omission, where relevant skills are discarded because they exceed the context budget, and attentional interference, where irrelevant skills dilute LLM's reasoning capacity. Existing methods either present all skills indiscriminately or rely on coarse retrieval strategies that fail to respect the tight coupling between skill applicability and the structural preconditions of individual cases. This causes a \emph{scalability routing problem}: how to identify the minimal yet sufficient subset of skills for a given case while strictly adhering to context constraints.

To address these three challenges, we propose KDASO. Rather than treating skill generation, refinement, and selection as isolated engineering components, KDASO frames them as interdependent solutions to the core problems of coverage, alignment, and scalability. Specifically, to close the skill coverage gap under distribution shift, we introduce \textbf{Neighborhood-Conditioned Skill Induction}, which leverages local manifold structure to identify representative case neighborhoods for skill generation, balancing the specificity of single-case induction with the feasibility of full-set processing. To bridge the semantic execution gap, we develop \textbf{Execution-Aligned Skill Refinement}, which analyzes discrepancies between LLM execution trajectories and expert reasoning paths, then restructures skill formulations to improve trajectory stability. To solve the scalability routing problem, we design \textbf{Context-Budgeted Skill Routing}, which trains lightweight applicability classifiers to dynamically route the most relevant skills for each case while guaranteeing context budget compliance.

Our contributions are summarized as follows:
\begin{itemize}
    \item \textbf{Problem.} We formalize three fundamental challenges in skill-based liability adjudication, including skill coverage under distribution shift, skill execution alignment and skill scalability under context budget.
    \item \textbf{Analysis.} We conduct extensive empirical analysis and identify that LLMs frequently fail to adapt to dynamic-evolving scenarios, deviate from intended logic when skills are ambiguous and disregard most skills when presented with extensive ones.
    \item \textbf{Method.} We propose KDASO, a unified framework that addresses these challenges through neighborhood-conditioned skill induction, execution-aligned skill refinement, and context-budgeted skill routing, enabling continuous evolution of domain-specific skills.
    \item \textbf{Evaluation.} We conduct comprehensive experiments across three real-world scenarios, demonstrating that KDASO achieves state-of-the-art performance. We further improve performance by 8.96\% when transferring skills to concept-similar scenarios and maintains consistent gains through continuous optimization.
\end{itemize}

\begin{figure*}[t]
\centering
\begin{minipage}[t]{0.45\linewidth}
    \centering
    \includegraphics[width=0.9\linewidth]{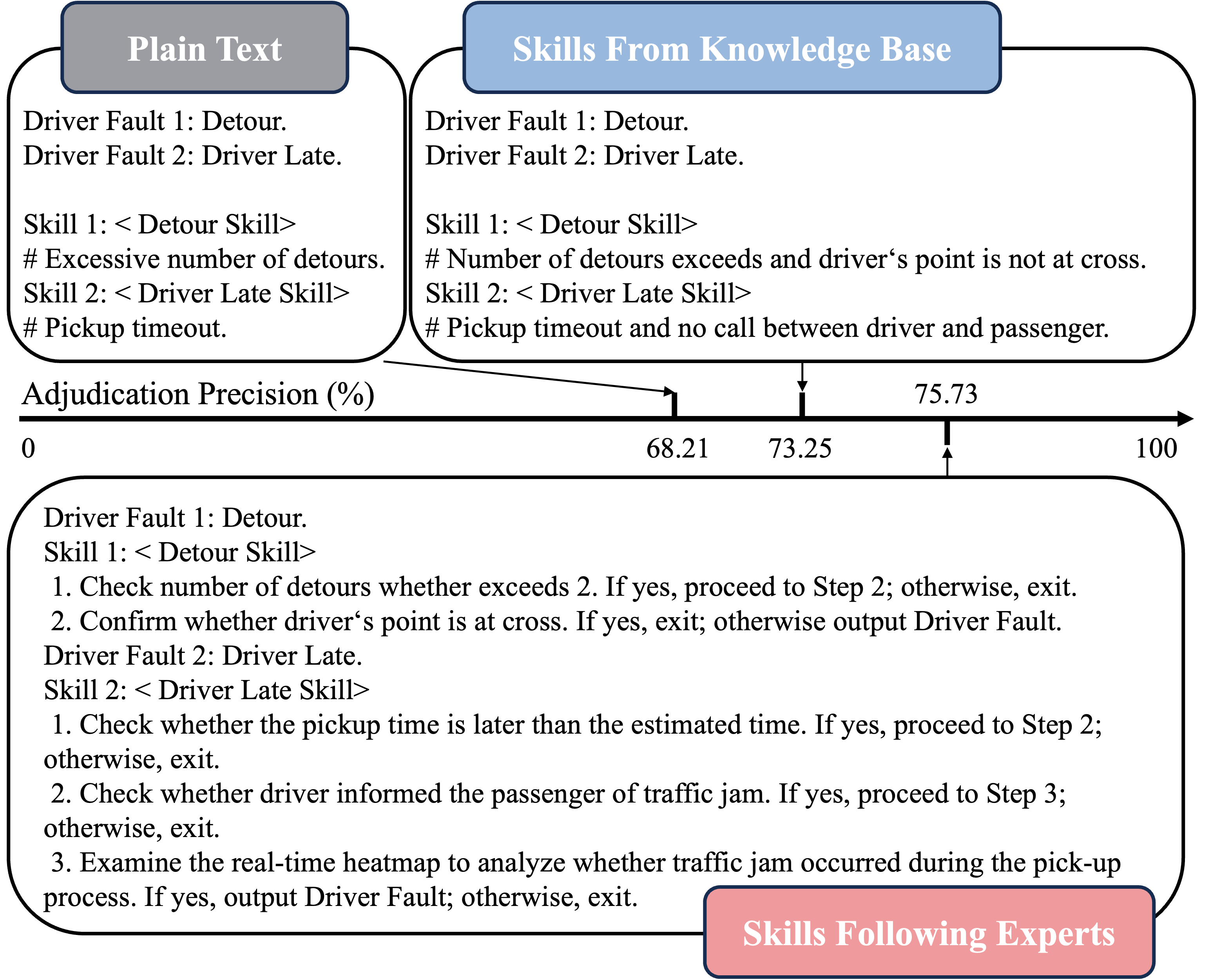}
    \caption{The impact of different logical description quality of skills on model performance. We select plain text, knowledge base-derived, and following experts to represent skill description quality levels.}
    \label{ob2}
\end{minipage}
\hspace{0.04\linewidth}
\begin{minipage}[t]{0.45\linewidth}
    \centering
    \includegraphics[width=0.95\linewidth, trim=0cm 0cm 1.5cm 0cm, clip]{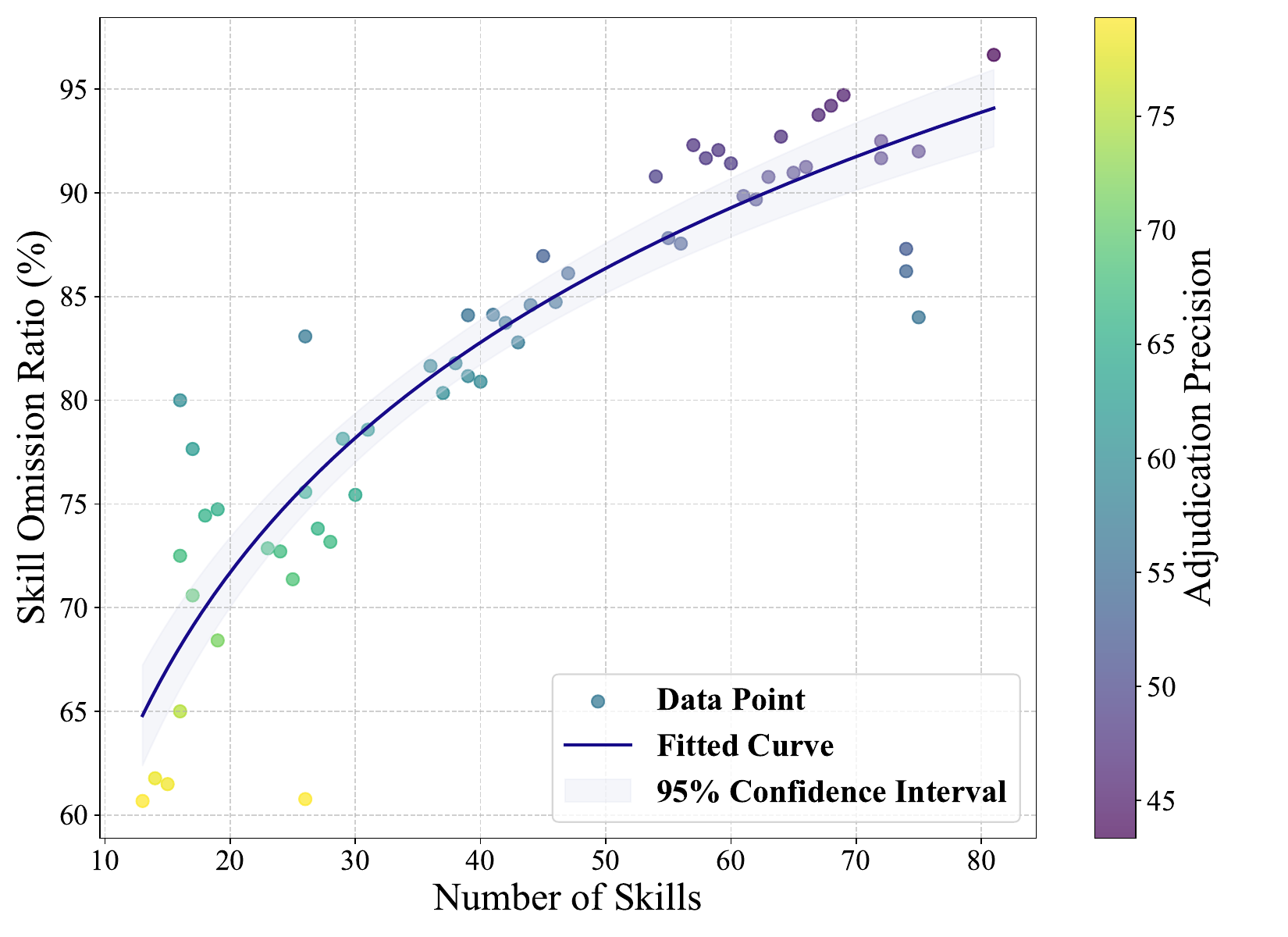}
    \caption{Number of skills as input v.s. skill omission ratio with adjudication precision evaluated on LLMs. Notably, the skill omission ratio correlates positively with input skill count, with $R^2$ = 0.82.}
    \label{ob1}
\end{minipage}
\vspace{-2pt}
\end{figure*}

\section{Related Work}
\subsection{Liable-Adjudicated General-Purpose Algorithms}
Previous research on liable-adjudicated general-purpose algorithms has employed two primary methodological strands. Logic-reasoning approaches formalize domain-specific rules and event descriptions into structured representations~\cite{peterson1977petri,cheng2025realistic,lv2026can}, such as first-order predicate logic~\cite{colmerauer1990introduction}, subsequently performing explicit deductive inference. Semantic-analysis methods utilize pretrained language models~\cite{aftan2023survey} and task-specific encoders~\cite{sun2025a} to map both rules and evidence into continuous vector spaces, determining liability through semantic similarity~\cite{chandrapati2024descriptive} or classification~\cite{liu2025simple}. Recently, LLMs have been developed to enhance performance in liability adjudication. Furthermore, RAG-based LLMs focus on enhancing the quality of rules. SIM-RAG~\cite{yang2025knowing} simultaneously performs question-answering and question generation. NodeRAG~\cite{xu2025noderag} employs rules mined from knowledge graphs to guide both retrieval and generation. However, these liable-adjudicated general-purpose algorithms still fail to adapt to dynamic-evolving scenarios, deviate from intended logic when rules are ambiguous, and disregard rules when presented with extensive ones. Hence, to address these three limitations, we design dedicated modules tailored to each challenge: a dynamic adaptation module that enables the framework to keep pace with evolving scenarios, an ambiguity-resolution module that preserves logical consistency when rules are underspecified, and a rule-prioritization module that ensures critical rules remain salient under conditions of rule overload.

\subsection{Liable-Adjudicated Domain-Specific Models}
Liability adjudication tasks primarily encompass domains such as ride-hailing adjudication, legal liability determination and logical reasoning judgment~\cite{matarazzo2025survey}. In the ride-hailing domain, current methods remain at the stage of applying liable-adjudicated algorithms. Research in the law and logic domain has mainly concentrated on training a specific expert model. For the legal domain, ChatLaw~\cite{cui2024chatlaw} and LAWGPT~\cite{zhou2025lawgpt} fine-tune foundation LLMs on specially curated legal corpora to strengthen their legal reasoning abilities. In the field of logical reasoning, data-level integrations, such as AMR-LDA~\cite{bao2024abstract} and LReasoner~\cite{wang2022logic}, inject formal logic through structured data augmentation. Parameter-level enhancements, such as MERIt~\cite{jiao2022merit} and IDOL~\cite{xu2023idol}, strengthen foundational reasoning through logic-aware pretraining techniques. However, these liable-adjudicated domain-specific models are costly to train and cannot be directly transferred to other domains. To address this, our
KDASO not only enables rapid deployment across various domains but also facilitates direct rule
transfer between concept-similar domains.

\section{Formulation and Challenges}
\subsection{Problem Formulation}
In this section, we formally define the liability adjudication task, which requires a reasoner \( f: (x, \mathcal{S}) \rightarrow y \) to determine liability based on input data \(x\) and a set of domain-specific skills \(\mathcal{S}\). A skill $s$ is a tuple $s = (\mathcal{C}_s, \mathcal{A}_s, \mathcal{T}_s)$, where $\mathcal{C}_s$ denotes the precondition set that determines skill triggering, $\mathcal{A}_s$ denotes the adjudication logic producing the output, and $\mathcal{T}_s$ denotes the structural template governing execution flow. The output \(y\) consists of a reasoning trajectory \(y_T\) and a final liability label \(y_L\). For a test input \(x_{\text{test}}\), the reasoner applies the current skills \(\mathcal{S}\) to generate \({y_{\text{test}}}_T\) and predict \({y_{\text{test}}}_L\).

\subsection{Empirical Analysis}
\label{3.2}
We identify three observations that also serve as key challenges explaining why current agents exhibit suboptimal performance on liability adjudication tasks. We evaluate QwQ-32B on ride-hailing liability adjudication scenarios using precision as the evaluation metric.

\textbf{\underline{Observation 1: }Static and lagging skills fail to adapt to dynamic-evolving scenarios.}  

In practical liability adjudication scenarios, current skills in domain-specific knowledge bases are largely derived from expert-based empirical generalizations of historical cases~\cite{bharadwaj2017creation}, exhibiting a notable temporal lag. As the domain scale continuously expands, novel patterns continually emerge, outpacing the rate at which human-curated skills can be formalized and integrated. For instance, in ride-hailing scenarios, a recently emerged "virtual displacement" fraud involves ride-hailing drivers using portable GPS spoofing devices to instantly "jump" their reported location several kilometers after accepting an order, falsely claiming timely arrival and blaming passengers for no-shows to illicitly obtain platform compensation. As such behavior leaves no trace in historical data because this hardware-based manipulation has only become feasible with the recent proliferation of high-precision dual-mode GPS, existing skills cannot recognize this novel, technology-enabled malicious pattern, therefore deeming the driver not liable. This omission prevents models from accessing core inferential grounds when confronting complex or emerging cases. Although current skills adequately cover most high-frequency situations, they lack the agility to respond to dynamic operational realities~\cite{cheng2025tabfsbench}. This underscores the need for a skill-induction framework that enables continuous extraction of new skills from the latest adjudications to support real-time evolution of skills.

\textbf{\underline{Observation 2: }Logic and description quality of skills determine skill executability by LLMs.}  

We evaluate LLMs under varying logic and description quality of skills to analyze how these affect performance on liability adjudication tasks. As shown in Figure~\ref{ob2}, the higher the quality of skills and the closer their formulation aligns with the authentic reasoning trajectories of human experts, the more effectively LLMs can comprehend and utilize them, leading to improved adjudication performance. Specifically, current LLMs exhibit limited capabilities in semantic generalization and handling ambiguous decision boundaries~\cite{li2025knowledge}, relying on literal pattern matching rather than deep inferential reasoning~\cite{pmlr-v267-modarressi25a}. However, a significant portion of adjudication rationale often resides in submerged logic, which is informal, unstructured experiential knowledge embedded in human decision-making, which has frequently not yet been codified or integrated into current skills. For instance, when a passenger claims a detour occurred, an experienced adjudicator checks whether the number of detours is excessive and further considers external factors, such as whether the pickup point is at an intersection or if traffic congestion exists. However, current skills only vaguely state "check whether the number of detours is excessive and the driver's point is at a cross" without specifying the required evidence. This ambiguity hinders LLMs from correctly analysing the appropriate evidence. Given the necessity of producing detailed, high-quality skill logics and descriptions, an automated execution-aligned skill refinement mechanism is essential to enhance the executability by LLMs.
\vspace{20pt} 

\begin{figure*}[t]
\begin{center}
\centerline{\includegraphics[width=\linewidth, trim=1cm 6.8cm 1cm 0cm, clip]{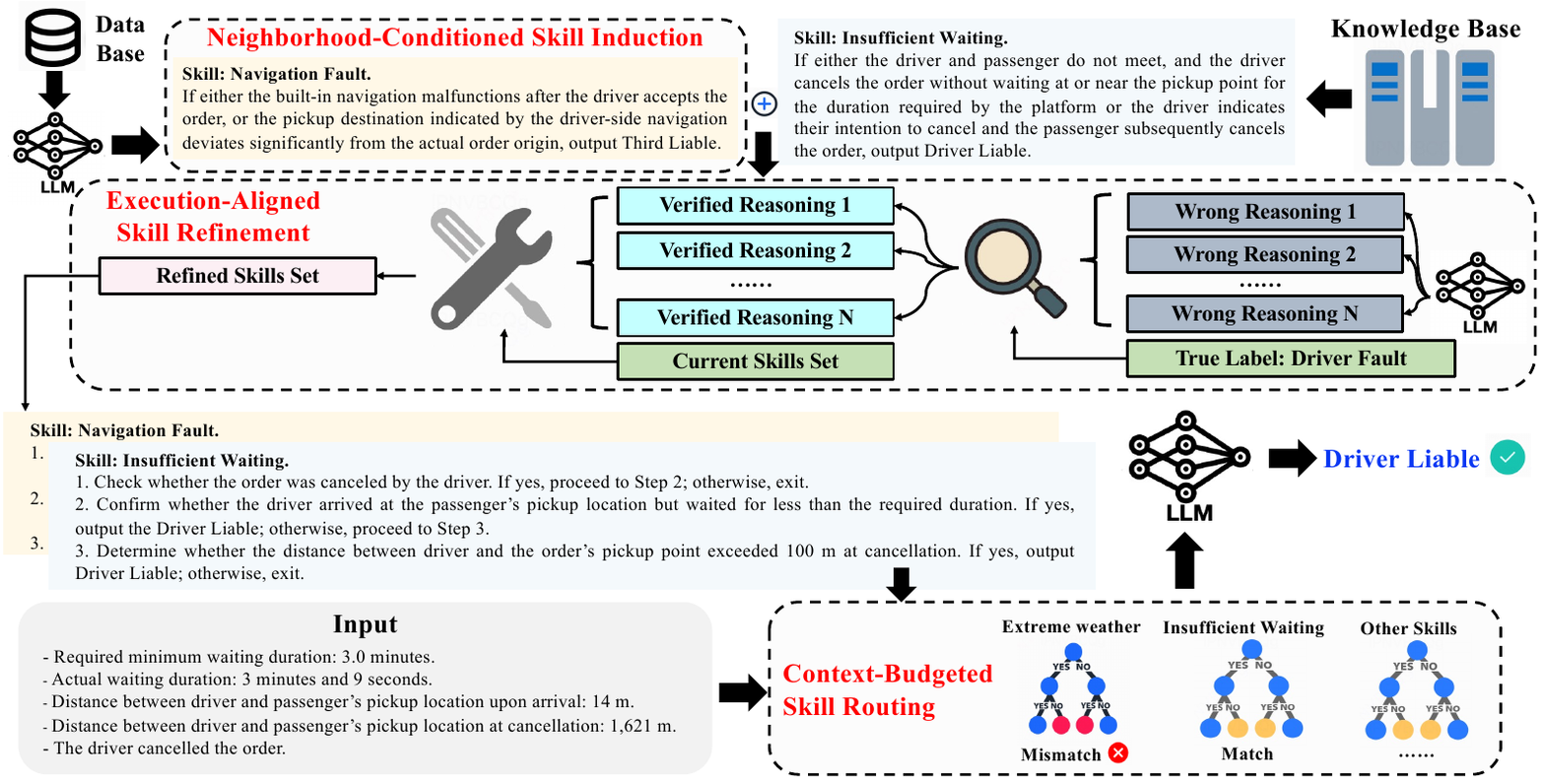}}
\caption{The overall illustration of KDASO framework.}
\label{pipeline}
\end{center}
\vspace{-0.2 in}
\end{figure*}

\textbf{\underline{Observation 3: }Skill omission ratio correlates positively with input skill count.}  

We conduct a quantitative comparison between the number of skills provided as input to LLMs and the ratio of skills overlooked during testing. We define a skill as overlooked if the LLM fails to mention or analyze that particular skill when generating its reasoning output. The length of a single skill ranges from 100 to 200 tokens, and we add skills incrementally in descending order of usage frequency. As illustrated in Figure~\ref{ob1}, a common phenomenon of skill omission emerges in multi-skill reasoning by LLMs: the ratio of omissions exhibits a positive correlation with the total number of skills provided as input. Our experiments further reveal a performance decay that accompanies the increasing skill omission ratio. The root cause of such omission lies in the inherent selectivity of the Transformer self-attention mechanism~\cite{vaswani2017attention}, which allocates attention weights unevenly across input skills, thereby causing certain skills to be insufficiently activated or even excluded from the reasoning process altogether~\cite{shaw2018self, peng2019self}. This bias introduces heightened risks of misjudgment in high-skill-density liability adjudication scenarios. Consequently, a context-budgeted skill routing mechanism is essential to extract critical and input-relevant skills prior to adjudication.

\section{Methodology}
\label{sec:methodology}

To address the challenges identified in Section~\ref{3.2}, we propose KDASO, an adaptive skill-based adjudication framework comprising three synergistic modules: \textit{Neighborhood-Conditioned Skill Induction}, \textit{Execution-Aligned Skill Refinement}, and \textit{Context-Budgeted Skill Routing}. The framework operates on a cumulative adjudication dataset \(\mathcal{D} = \{(x^j, y^j)\}_{j=1}^{N}\). A skill \(s\) is executed on input \(x\) as \(f(x, s) \rightarrow y\) whenever \(x\) satisfies the skill's preconditions, and the skill library \(\mathcal{S}\) collects all validated skills. Figure~\ref{pipeline} illustrates the framework architecture.

\subsection{Neighborhood-Conditioned Skill Induction}
\label{sec:skill_induction}

Skills induced from a single case $(x^i, y^i)$ risk capturing idiosyncratic patterns that fail to generalize to perturbed inputs, while induction over the full dataset $\mathcal{D}_{\text{new}}$ is intractable due to case heterogeneity and context length exceeding the LLM window limit $L_{\text{ctx}}$. We address this tension via neighborhood-conditioned induction that leverages local manifold structure.

Given a new adjudication batch $\mathcal{D}_{\text{new}} = \{(x^i, y^i)\}_{i=1}^{N}$ with $y^i = (y_L^i, y_T^i)$, we embed instances via $\phi: \mathcal{X} \rightarrow \mathbb{R}^d$ and compute cosine similarities $\text{sim}(x^i, x^j) = \langle \phi(x^i), \phi(x^j) \rangle / (\|\phi(x^i)\| \cdot \|\phi(x^j)\|)$. The top-$k$ most similar instances form the neighborhood $\mathcal{N}_k(x^i) = \{(x^j, y^j)\}_{j=1}^{k}$, which constitutes a local chart on the data manifold providing sufficient but non-redundant evidence for skill generation. LLM then induces a new skill$s_{\text{new}}^i$ via $\mathcal{N}_k(x^i)$. This design avoids overfitting to instance noise while keeping input complexity $\ll L_{\text{ctx}}$, isolating the induction from semantically distant cases.

Post-induction, each $s_{\text{new}}^i$ undergoes validation. Its support set $\mathcal{S}_{\text{supp}}(s_{\text{new}}^i) = \{(x, y) \in \mathcal{D}_{\text{new}} \mid f(x, s_{\text{new}}^i) \rightarrow y\}$ must be non-empty. Otherwise the skill is hallucinated and discarded. For conflict detection with existing $s_{\text{old}} \in \mathcal{S}$, if both skills trigger on some $(x', y')$ but yield conflicting outputs, we compare their empirical accuracies on the overlapping domain and discard the inferior skill, or merge them via LLM synthesis if both fail. We finally yield the augmented skill library $\mathcal{S}'$.

\subsection{Execution-Aligned Skill Refinement}
\label{sec:skill_refinement}

A skill logically comprehensible to human experts may induce inconsistent LLM execution trajectories due to divergent priors or ambiguous natural language formulations. Let $\mathcal{H}(x, s)$ denote the human expert trajectory and $\mathcal{L}(x, s; \theta)$ the LLM trajectory under weights $\theta$. Even when both converge to $y_L$, their execution paths may diverge: $\mathbb{P}_{\theta}[\mathcal{L}(x, s; \theta) \neq \mathcal{H}(x, s) \mid f(x, s) \rightarrow y] > 0$.

For each $s^i \in \mathcal{S}'$, retrieve an anchor case $(x_{s^i}, y_{s^i})$ where $f(x_{s^i}, s^i) \rightarrow y_{s^i}$. Perform $G$ stochastic forward passes to sample trajectories $\{y_{s^i}^j\}_{j=1}^{G} \sim \mathbb{P}_{\theta}(\cdot \mid x_{s^i}, s^i)$ with $y_{s^i}^j = (y_L^j, y_T^j)$, and measure similarity to the human reference as $c_i^j = \text{sim}(\phi(y_{s^i T}^j), \phi(y_{s^i T})) \in [-1,1]$. The misalignment set $\mathcal{M}_i = \{(y_{s^i T}^j, c_i^j) \mid c_i^j < \tau\}$ captures trajectories with substantial reasoning discrepancy, and the instability rate $\rho(s^i) = |\mathcal{M}_i| / G$ quantifies execution variance. Skills with $\rho(s^i) > \rho_{\max}$ are flagged for refinement via the module.

The refinement process feeds $(x_{s^i}, y_{s^i}, s^i, \mathcal{M}_i)$ into the LLM, which reconstructs the expert reasoning path, examines misaligned trajectories in ascending order of $c_i^j$ to diagnose failure modes (e.g., divergent subgoal ordering or omitted preconditions), and synthesizes a revised skill $s^{i'}$ with explicit execution scaffolding following the chain-of-thought template~\cite{wei2022chain}. This iterates until all $G$ trajectories satisfy $c_i^j \geq \tau$ or a maximum iteration budget is reached, yielding the refined library $\mathcal{S}''$.

\subsection{Context-Budgeted Skill Routing}
\label{sec:skill_routing}

As $|\mathcal{S}|$ grows, feeding the entire library alongside $x_{\text{test}}$ exceeds the context window: $\ell(x_{\text{test}}) + \sum_{s \in \mathcal{S}} \ell(s) > L_{\text{ctx}}$, causing skill omission and attentional interference from irrelevant skills. We propose a routing module that selects a minimal sufficient subset $\mathcal{S}_{\text{final}} \subseteq \mathcal{S}''$ prior to LLM invocation. For each $s^i \in \mathcal{S}''$, construct a skill-specific dataset $\mathcal{D}_{s^i} = \{(x^j, \hat{y}_i^j)\}_{j=1}^{N}$ where $\hat{y}_i^j = \mathbb{I}[f(x^j, s^i) \rightarrow y^j] \in \{0,1\}$ indicates whether $s^i$ correctly triggers on $x^j$. Train a lightweight classifier $m^i: \mathcal{X} \rightarrow \{0,1\}$ via maximum likelihood on $\mathcal{D}_{s^i}$ using XGBoost~\cite{10.11452939672.2939785}.

At test time, given $x_{\text{test}}$, the retrieval stage collects skills predicted as applicable: $\mathcal{S}_{\text{retrieve}} = \{s^i \in \mathcal{S}'' \mid m^i(x_{\text{test}}) = 1\}$. To mitigate classifier false negatives, augment with direct structural matches $\mathcal{S}_{\text{match}} = \{s^i \in \mathcal{S}'' \mid x_{\text{test}} \models s^i\}$, yielding $\mathcal{S}_{\text{final}} = \mathcal{S}_{\text{retrieve}} \cup \mathcal{S}_{\text{match}}$.

To respect the context budget, compute the maximum storable skills $k_{\max}$ such that $\ell(x_{\text{test}}) + \sum_{j=1}^{k_{\max}} \ell(s^{\sigma(j)}) \leq L_{\text{ctx}}$ where skills are sorted by priority combining classifier confidence and direct match signals. If $|\mathcal{S}_{\text{final}}| > k_{\max}$, truncate to the top-$k_{\max}$ skills. The final adjudication is $y_{\text{test}} = f(x_{\text{test}}, \mathcal{S}_{\text{final}}^{\text{truncate}})$, ensuring completeness and feasibility.

\paragraph{Global Objective.} Over temporal horizon $T$, the framework maximizes cumulative trajectory alignment while controlling library growth:
\begin{equation}
\max_{\{\mathcal{S}^{(t)}\}_{t=1}^{T}} \sum_{t=1}^{T} \left( \mathbb{E}_{(x,y) \sim \mathcal{D}^{(t)}} \left[ \text{sim}\left( \phi(\mathcal{L}(x, \mathcal{S}_{\text{route}}^{(t)}(x); \theta)), \phi(\mathcal{H}(x)) \right) \right] - \gamma \cdot |\mathcal{S}^{(t)}| \right)
\end{equation}
where $\mathcal{S}_{\text{route}}^{(t)}(x)$ denotes the routed subset at epoch $t$ and $\gamma$ penalizes excessive growth. This closed-loop KDASO ensures $\mathcal{S}$ remains accurate, compact and adaptable to evolving adjudication scenarios.

\section{Experiment Setup}
\subsection{Datasets}
We list the original datasets and skills used in this paper. Detailed information on all datasets and construction methods on LegalBench can be found in Appendix~\ref{exp:datasets}. 

\textbf{Ride-hailing Adjudication.}
We select five closed-source datasets from DiDi as ride-hailing adjudication tasks. Adjudication tasks in DiDi ride-hailing scenarios can be broadly categorized into Cancel and Fee. Cancel-related scenarios involve the cancellation of an order by either the driver or the passenger prior to driver pickup, whereas Fee-related scenarios pertain to disputes between drivers and passengers regarding the actual fee charged during the trip. From the Cancel category, we select four representative subtasks: 1) \textit{Cancel–Appeal (CA)}: driver appeals after order cancellation; 2) \textit{Cancel–Passenger cancel (CPC)}: order canceled by the passenger; 3) \textit{Cancel–Driver cancel (CDC)}: order canceled by the driver; 4) \textit{Cancel–Reservation (CR)}: pre-scheduled reservation orders canceled before the scheduled time takes effect. Within the Fee category, we focus on \textit{Fee–Detour (FD)}, which concerns fee increases resulting from actual driving distances significantly exceeding the estimated distances. Notably, we evaluate CPC and CDC only in Section~\ref{5.3}. We select relatively hard cases into ride-hailing datasets. We adopt adjudication standards in DiDi for ride-hailing skills.

\textbf{Legal Liability Determination.}
In the legal domain, we adopt \textit{Issue Topic Identification (LA-ITI)} and \textit{Charge Prediction (LA-CP)} from LawBench~\cite{fei2024lawbench} and manually construct \textit{OPP-115 (LE-OPP), Unfair (LE-UN)} and \textit{Learned Hand (LE-LH)} based on LegalBench~\cite{guha2023LegalBench} as legal liability determination tasks. We define legal skills from real-world legal doctrines and statutes.

\textbf{Logical Reasoning Judgment.}
We choose \textit{ReClor}~\cite{yu2020reclor} and \textit{AR-LSAT}~\cite{zhong2021ar} as logical reasoning judgment tasks. We treat training datasets from ReClor and AR-LSAT as logical skills.

\begin{table*}[t]
\centering
\caption{Score comparison of \textsc{KDASO} against liability adjudication general-purpose algorithms and open-source general LLMs across three tasks from ride-hailing adjudication. The best is in \textbf{bold} and the second best is \underline{underlined} among non-proprietary models.}
\label{complete}
\begin{sc}
\resizebox{\textwidth}{!}{
\begin{tabular}{ccccccccccccc}
\toprule
\toprule
\multirow{2}{*}{Model}$^\dagger$  & \multicolumn{4}{c}{CA} & \multicolumn{4}{c}{CR} & \multicolumn{4}{c}{FD} \\
\cmidrule(lr){2-5} \cmidrule(lr){6-9} \cmidrule(lr){10-13}
 & O-Precision & O-Recall & B-Precision & B-Recall & O-Precision & O-Recall & B-Precision & B-Recall & O-Precision & O-Recall & B-Precision & B-Recall \\
\midrule
NL2FOL & 68.21\%$\pm$0.54\% & 62.13\%$\pm$1.87\% & 41.84\%$\pm$0.32\% & 23.56\%$\pm$0.45\% & 62.46\%$\pm$0.73\% & 50.66\%$\pm$0.68\% & 73.96\%$\pm$0.91\% & 30.31\%$\pm$0.29\% & 77.17\%$\pm$0.64\% & 60.85\%$\pm$0.77\% & 9.72\%$\pm$0.19\% & 43.75\%$\pm$0.58\% \\
NeoBERT & 57.98\%$\pm$1.62\% & 11.02\%$\pm$0.21\% & 52.94\%$\pm$0.48\% & 36.21\%$\pm$0.56\% & 63.95\%$\pm$0.71\% & 24.13\%$\pm$0.33\% & 56.30\%$\pm$0.59\% & 35.66\%$\pm$0.47\% & 71.13\%$\pm$0.83\% & 9.61\%$\pm$0.18\% & \textbf{14.43\%$\pm$0.26\%} & 17.50\%$\pm$0.51\% \\
ModernBERT & 63.44\%$\pm$0.75\% & 18.85\%$\pm$0.28\% & 47.31\%$\pm$0.48\% & \underline{46.57\%$\pm$0.69\%} & 59.02\%$\pm$0.66\% & 16.73\%$\pm$0.04\% & 50.95\%$\pm$0.52\% & 24.24\%$\pm$0.35\% & 70.30\%$\pm$0.79\% & 9.89\%$\pm$0.20\% & 13.86\%$\pm$0.23\% & 17.50\%$\pm$1.31\% \\
SIM-RAG & 71.31\%$\pm$0.88\% & 59.16\%$\pm$0.74\% & 57.78\%$\pm$0.65\% & 29.89\%$\pm$0.41\% & 58.43\%$\pm$0.92\% & 15.48\%$\pm$0.42\% & 51.24\%$\pm$0.45\% & 23.75\%$\pm$0.34\% & 74.27\%$\pm$0.85\% & 37.46\%$\pm$0.51\% & 12.36\%$\pm$0.21\% & 23.75\%$\pm$0.34\% \\
NodeRAG & 72.24\%$\pm$2.89\% & 61.70\%$\pm$0.76\% & 56.82\%$\pm$0.63\% & 28.74\%$\pm$0.39\% & 61.66\%$\pm$0.90\% & 31.21\%$\pm$0.53\% & 50.02\%$\pm$1.33\% & 24.63\%$\pm$0.36\% & 74.91\%$\pm$0.86\% & 45.23\%$\pm$1.89\% & \underline{14.22\%$\pm$0.25\%} & 25.49\%$\pm$0.37\% \\
Qwen3-8B & 68.08\%$\pm$1.82\% & 74.28\%$\pm$0.90\% & 34.50\%$\pm$0.46\% & 4.02\%$\pm$0.11\% & 28.56\%$\pm$0.40\% & \textbf{85.44\%$\pm$0.95\%} & 52.15\%$\pm$0.77\% & 29.01\%$\pm$0.20\% & \underline{78.06\%$\pm$2.92\%} & 60.44\%$\pm$1.83\% & 10.03\%$\pm$0.20\% & 40.00\%$\pm$0.55\% \\
QwQ-32B & \underline{72.49\%$\pm$0.89\%} & 62.30\%$\pm$0.97\% & 57.78\%$\pm$0.65\% & 29.88\%$\pm$0.41\% & 38.78\%$\pm$0.52\% & \underline{85.32\%$\pm$0.94\%} & 55.17\%$\pm$0.61\% & 54.24\%$\pm$0.60\% & 76.55\%$\pm$0.87\% & \underline{63.65\%$\pm$0.78\%} & 9.92\%$\pm$0.49\% & \underline{45.00\%$\pm$0.29\%} \\
KDASO\textsubscript{Qwen3-8B} & 69.56\%$\pm$0.83\% & \textbf{92.01\%$\pm$0.98\%} & \textbf{69.79\%$\pm$0.84\%} & 38.50\%$\pm$0.53\% & \underline{74.41\%$\pm$0.85\%} & 79.53\%$\pm$0.93\% & \textbf{83.76\%$\pm$0.96\%} & \underline{71.49\%$\pm$0.86\%} & 77.99\%$\pm$0.89\% & 63.65\%$\pm$0.84\% & 10.03\%$\pm$0.20\% & 43.75\%$\pm$0.58\% \\
KDASO\textsubscript{QwQ-32B} & \textbf{73.42\%$\pm$0.90\%} & \underline{87.38\%$\pm$0.97\%} & \underline{69.17\%$\pm$0.89\%} & \textbf{47.70\%$\pm$0.60\%} & \textbf{80.39\%$\pm$0.94\%} & 79.09\%$\pm$0.92\% & \underline{81.18\%$\pm$0.95\%} & \textbf{75.67\%$\pm$0.87\%} & \textbf{79.07\%$\pm$0.91\%} & \textbf{66.29\%$\pm$0.81\%} & 10.11\%$\pm$0.20\% & \textbf{47.50\%$\pm$0.60\%} \\
\bottomrule
\bottomrule
\end{tabular}}
\end{sc}
\begin{tablenotes}
\item \footnotesize $^\dagger$ Proprietary LLMs and DiDi’s internally liable-adjudicated domain-specific models are omitted due to data protection protocols and proprietary company information. Correspondingly, SIM-RAG and NodeRAG, initially designed with Proprietary LLMs dependencies, are reconfigured to employ the locally instantiated QwQ-32B as an alternative.
\end{tablenotes}
\vspace{-0.1in}
\end{table*}

\subsection{Models}
We compare KDASO against four categories of models: proprietary LLMs, liable-adjudicated general-purpose algorithms, liable-adjudicated domain-specific models and open-source general LLMs. We provide model details in Appendix~\ref{exp:models}. Notably, we only evaluate liable-adjudicated domain-specific models in their corresponding domains. \textbf{Proprietary LLMs.} We evaluate Gemini 3 Pro and GPT-4o~\cite{hurst2024gpt} as representatives of proprietary LLMs. \textbf{Liable-Adjudicated General-Purpose Algorithms.} We choose NL2FOL~\cite{lalwani2025autoformalizing}, NeoBERT~\cite{le2025neobert} and ModernBERT~\cite{warner2025smarter}, prompt-based SIM-RAG~\cite{yang2025knowing} and NodeRAG~\cite{xu2025noderag} which requires training. \textbf{Liable-Adjudicated Domain-Specific Models.} We choose Lawyer-LLAMA~\cite{huang2023lawyer}, LexiLaw~\cite{LexiLaw} and ChatLaw~\cite{cui2024chatlaw} for the law domain, and select MERIt~\cite{jiao2022merit}, LReasoner Ensemble~\cite{wang2022logic}, IDOL~\cite{xu2023idol} and AMR-LDA~\cite{bao2024abstract} for the logic domain. \textbf{Open-Source General LLMs.} We select Deepseek-R1~\cite{guo2025deepseek}, Qwen3-8B~\cite{yang2025qwen3} and QwQ-32B as the newest representatives of open-source general LLMs for evaluation and the backbones for KDASO.

\subsection{Metrics}
In ride-hailing adjudication, we categorize liability into two distinct levels: ordinary liability (O) and bad faith misconduct (B), following DiDi's standards. For each of these two levels, we separately evaluate model performance using precision and recall. For legal and logic datasets, we adopt the evaluation metrics officially recommended by their corresponding benchmarks. Detailed descriptions of all evaluation metrics are provided in Appendix~\ref{exp:metrics}.

\section{Experiment Results}
In this section, we evaluate the performance of KDASO in scenarios on ride-hailing adjudication, legal liability determination and logical reasoning judgment. Furthermore, we directly apply generated skills to other concept-similar scenarios to test transferability, and continuously optimize them to assess their stability. Finally, we conduct an ablation study to empirically demonstrate the effectiveness of three core components in KDASO. Extended experiments are shown in Appendix~\ref{G}\~~\ref{symbol} and the case study is in Appendix~\ref{case}.
\label{5.2}

\begin{table*}[t]
    \centering
    \scriptsize

    \begin{minipage}[t]{0.48\textwidth}
        \captionof{table}{KDASO Score against compared models across five tasks from legal liability determination. The best is in \textbf{bold} and the second best is \underline{underlined} among non-proprietary models.}
        \label{tab:law}
        \setlength{\tabcolsep}{3pt}
        \begin{tabular}{@{}l|ccccc@{}}
            \toprule\toprule
            Model & LA-ITI & LA-CP & LE-OPP & LE-UN & LE-LH  \\
            \midrule
            \multicolumn{6}{@{}c@{}}{\cellcolor{prollm}Proprietary LLMs} \\
            \midrule
            \rowcolor{prollm}
            Gemini 3 Pro  & 39.8  & 62.9 & 82.9 & 85.8 & 83.2 \\
            \rowcolor{prollm}
            GPT-4o        & 44.8  & 52.9 & 77.0 & 87.2 & 76.6 \\
            \midrule
            \multicolumn{6}{@{}c@{}}{\cellcolor{lrsa}Liable-Adjudicated General-Purpose Algorithms} \\
            \midrule
            \rowcolor{lrsa}
            NL2FOL &34.0 & 53.8 & 69.8 & 83.5 & 74.3 \\
            \rowcolor{lrsa}
            NeoBERT     &38.9 & 51.7 & 71.7 & 83.0 & 75.0 \\
            \rowcolor{lrsa}
            ModernBERT  &33.4 & 53.7 & 68.8 & 83.7 & 74.5 \\
            \rowcolor{lrsa}
            SIM-RAG     &34.2 & 53.0 & 72.8 & 82.1 & 74.8 \\
            \rowcolor{lrsa}
            NodeRAG     &36.3 & 55.6 & 72.6 & 84.8 & 74.4 \\
            \midrule
            \multicolumn{6}{@{}c@{}}{\cellcolor{rag}Liable-Adjudicated Domain-Specific Models} \\
            \midrule
            \rowcolor{rag}
            Lawyer-LLAMA &4.4$^{\dagger}$ & 31.3 & 52.4 & 83.2 & 75.3 \\
            \rowcolor{rag}
            LexiLaw-6B &22.8 & 40.0 & 57.2 & 82.7 & 74.6 \\
            \rowcolor{rag}
            ChatLaw-13B &27.6 & 27.9 & 53.8 & 82.7 & 73.2\\
            \rowcolor{rag}
            ChatLaw-33B &19.8 & 12.7 & 56.1 & 84.3 & 71.7\\
            \midrule
            \multicolumn{6}{@{}c@{}}{\cellcolor{osgm}Open-Source General LLMs} \\
            \midrule
            \rowcolor{osgm}
            Deepseek-R1   & \textbf{44.2}  & \underline{61.0} & \underline{74.2} & 84.3 & 79.0\\
            \rowcolor{osgm}
            Qwen3-8B  & 33.6 & 53.6 & \underline{74.2} & 83.4 & 73.4\\
            \rowcolor{osgm}
            QwQ-32B   & 39.8 & 60.9 & \underline{74.2} & \underline{85.4} & \textbf{80.6}\\
            \midrule
            \rowcolor{KDASO}
            KDASO$_{\text{Qwen3}}$ & 39.6 & 56.4 & 73.8 & 83.4 & 76.1\\
            \rowcolor{KDASO}
            KDASO$_{\text{QwQ-32B}}$ & \underline{44.0} & \textbf{63.5} & \textbf{74.4} & \textbf{86.8} & \underline{80.4} \\
            \bottomrule\bottomrule
        \end{tabular}

        {\small $^{\dagger}$ Lawyer-LLAMA's low LA-ITI score is consistent with OpenCompass's website~\cite{buitrago2019open}.}
    \end{minipage}
    \hfill
    \begin{minipage}[t]{0.48\textwidth}
        \captionof{table}{KDASO Score against compared models across two tasks from logical reasoning judgment. The best is in \textbf{bold} and the second best is \underline{underlined} among non-proprietary models.}
        \label{tab:logic}
        \setlength{\tabcolsep}{7pt}
        \begin{tabular}{l|ccc|c}
            \toprule\toprule
            \multirow{2}{*}{Model}  & \multicolumn{3}{c|}{Reclor} & \multirow{2}{*}{AR-LSAT} \\
            \cmidrule(lr){2-4} 
             & Test & Test-E & Test-H & \\
            \midrule
            \multicolumn{5}{c}{\cellcolor{prollm}Proprietary LLMs} \\
            \midrule
            \rowcolor{prollm}
            Gemini 3 Pro  & 96.6 & 96.1 & 96.9 & 94.2 \\
            \rowcolor{prollm}
            GPT-4o        & 90.0 & 89.5 & 90.3 & 28.3$^{\dagger}$ \\
            \midrule
            \multicolumn{5}{c}{\cellcolor{lrsa}Liable-Adjudicated General-Purpose Algorithms} \\
            \midrule
            \rowcolor{lrsa}
            NL2FOL & 88.1 & 89.5 & 87.0 & 77.0\\
            \rowcolor{lrsa}
            NeoBERT  & 91.5 & 92.4 & \underline{90.8} & 80.1 \\
            \rowcolor{lrsa}
            ModernBERT & 85.5 & 88.2 & 83.4 & 78.9 \\
            \rowcolor{lrsa}
            SIM-RAG & 87.6 & 88.9 & 86.6 & 76.1 \\
            \rowcolor{lrsa}
            NodeRAG & 86.7 & 89.4 & 84.6 & 77.2 \\
            \midrule
            \multicolumn{5}{c}{\cellcolor{rag}Liable-Adjudicated Domain-Specific Models} \\
            \midrule
            \rowcolor{rag}
            MERIt & 79.3 & 85.2 & 74.6 & 73.0 \\
            \rowcolor{rag}
            LReasoner & 76.1 & 87.0 & 67.5 & 73.9 \\
            \rowcolor{rag}
            IDOL & 80.6 & 87.7 & 75.0 & 76.1 \\
            \rowcolor{rag}
            AMR-LDA & 90.2 & 91.6 & 89.1 & 77.0 \\
            \midrule
            \multicolumn{5}{c}{\cellcolor{osgm}Open-Source General LLMs}  \\
            \midrule
            \rowcolor{osgm}
            Deepseek-R1   & \textbf{96.0} & \textbf{96.1} & \textbf{95.9} & \textbf{95.2} \\
            \rowcolor{osgm}
            Qwen3-8B  & 87.3 & 87.9 & 86.8 & 77.0 \\
            \rowcolor{osgm}
            QwQ-32B   & 92.2 & \underline{94.1} & 90.7 & 83.9 \\
            \midrule
            \rowcolor{KDASO}
            KDASO$_{\text{Qwen3}}$ & 87.7 & 89.3 & 86.4 & 78.3 \\
            \rowcolor{KDASO}
            KDASO$_{\text{QwQ-32B}}$ & \underline{93.7} & 93.9 & \underline{93.6} & \underline{87.8} \\
            \bottomrule\bottomrule
        \end{tabular}
        
        {\small $^{\dagger}$ GPT-4o's low AR-LSAT score is consistent with~\cite{wang2025aqe}.}
    \end{minipage}
    \vspace{-0.2 in}
\end{table*}

\subsection{Ride-hailing Adjudication}
Table~\ref{complete} presents the performance of KDASO on three representative ride-hailing adjudication tasks (CA, CR and FD). Results demonstrate that KDASO outperforms the majority of compared models across the four evaluation metrics, exhibiting significant performance advantages. Specifically, compared to the strongest liable-adjudicated general-purpose algorithm, KDASO achieves average improvements of 6\% in O-Precision, 20\% in O-Recall, 5\% in B-Precision, and 15\% in B-Recall. Notably, following DiDi’s real-world adjudication standards, O-Precision is regarded as the most critical metric, as it directly reflects the fairness of the liable-adjudicated model and underpins user trust. KDASO delivers a remarkable 16\% absolute gain in O-Precision on CR, surpassing by a margin and exceeding practical expectations by closely aligning with the real-world requirement for high-precision, low-false-positive adjudication.

\subsection{Legal Liability Determination}
As shown in Table~\ref{tab:law}, KDASO outperforms existing models across various tasks from legal liability determination. KDASO, based on Qwen3-8B, also surpasses liable-adjudicated domain-specific models, demonstrating parameter efficiency. Notably, several models explicitly designed for legal tasks exhibit suboptimal performance on the adjudication task, suggesting that generic legal priors alone are insufficient for effective transfer to highly structured, skill-intensive scenarios. In contrast, KDASO bridges the gap between general semantic understanding and domain-specific skills through the integrated knowledge-data automated skill optimization. Moreover, on tasks such as LA-CP and LE-UN, KDASO achieves performance that even surpasses proprietary LLMs. 

\subsection{Logical Reasoning Judgment}
We further evaluate KDASO on ReClor and AR-LSAT. Our results have been submitted to the official leaderboard of Reclor, where KDASO currently ranks first. As shown in Table~\ref{tab:logic}, KDASO achieves a score gain of 3.5 over the previous best logic-specific model on Test set. This advantage is even more pronounced on Test-H set, which comprises particularly challenging logical reasoning instances, where KDASO demonstrates superior performance in complex logical reasoning scenarios. Moreover, on the AR-LSAT benchmark, KDASO surpasses QwQ-32B by 3.9\% improvement, further underscoring its effectiveness in complex, structured reasoning tasks.

\subsection{Transferability of Generated Skills by KDASO}
\label{5.3}
\begin{figure*}[t]
\centering
\begin{minipage}[t]{0.48\textwidth}
    \vspace{0pt}
    \centering
    \begin{minipage}[t]{\linewidth}
        \centering
        \vspace{0pt}
        \includegraphics[height=3.5cm, width=\linewidth, keepaspectratio, trim=1cm 0cm 0cm 0cm,  clip]{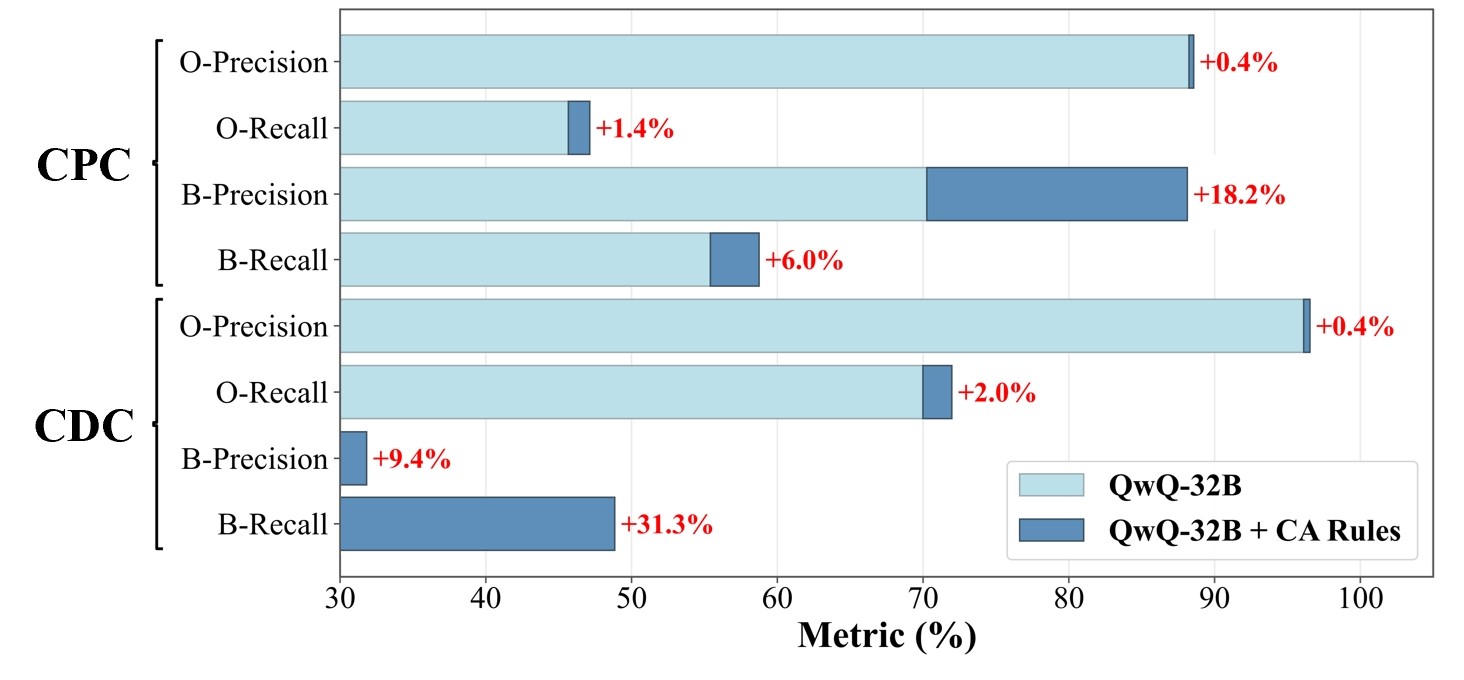}
        \caption{Model performance improvement in CPC and CDC based on skills generated by KDASO on CA. Red numbers to the right of bars indicate the improvement of performance.}
        \label{rq2}
    \end{minipage}
\end{minipage}\hfill
\begin{minipage}[t]{0.48\textwidth}
    \vspace{0pt}
    \centering
    \begin{minipage}[t]{\linewidth}
        \centering
        \vspace{0pt}
        \includegraphics[height=3.5cm, width=\linewidth, keepaspectratio, trim=0cm 0cm 0cm 0cm,  clip]{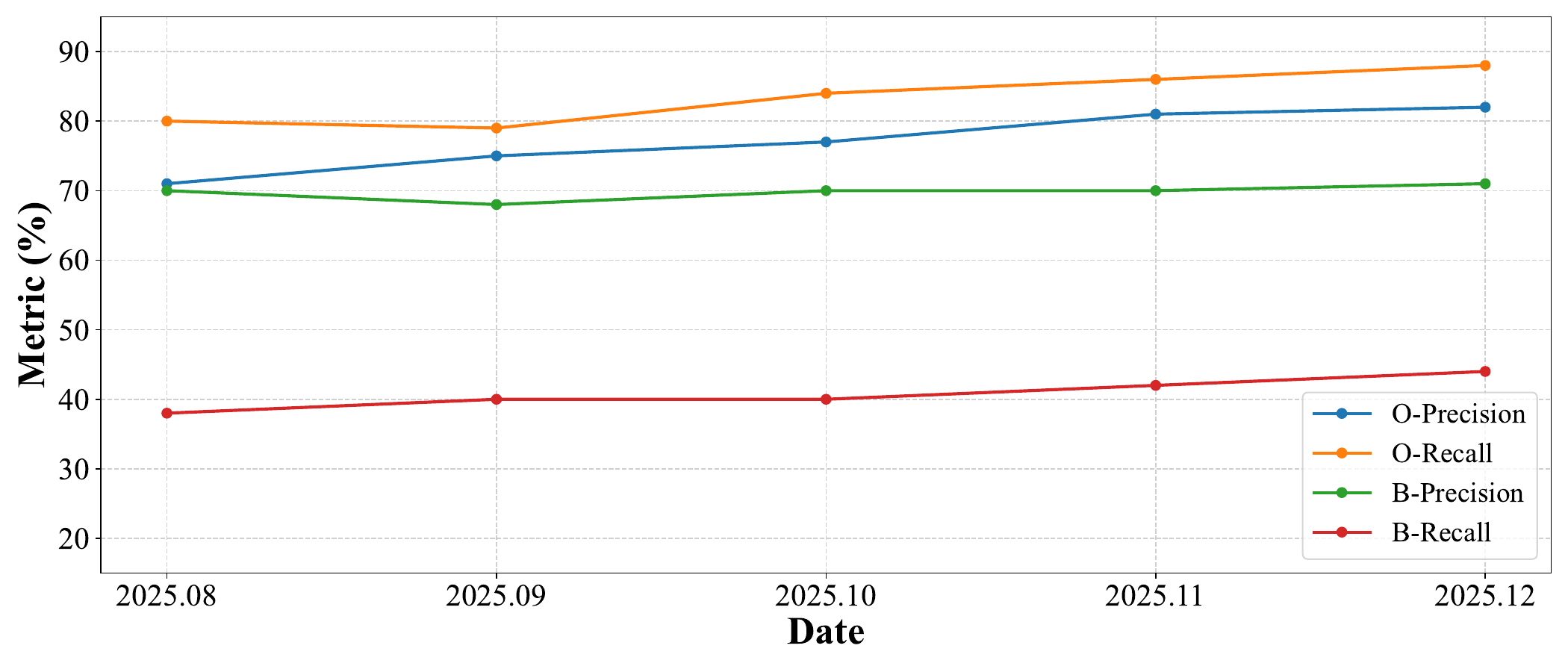}
        \vspace{-1pt}
        \caption{Comparison of KDASO’s performance over time. The $x$-axis corresponds to the date of generating-skill datasets.}
        \label{rq3}
    \end{minipage}
\end{minipage}
\vspace{-8pt}
\end{figure*}

To evaluate the transferability of the generated skills by KDASO, we transfer the skills generated on CA directly to two other tasks in ride-hailing adjudication, CPC and CDC, without any adaptation on KDASO. Specifically, we directly treat the skills generated from CA as skills in CPC and CDC and then evaluate QwQ-32B. As shown in Figure~\ref{rq2}, skills generated by KDASO in CA demonstrate strong transfer performance. While maintaining comparable O-Precision, skills achieve significant improvement in the other three metrics, especially in B-Precision and B-Recall. This result indicates that skills generated from CA by KDASO not only enhance adjudication performance within their original scenario but also substantially improve model effectiveness in semantically related downstream scenarios such as CPC and CDC. To date, 40\% of skills generated by KDASO have been successfully validated by DiDi team as novel yet consistent with existing policies and integrated into the skill base, providing empirical validation of their practical utility.

\begin{table}[t]
\centering
\caption{Ablation study of KDASO on CA.}
\vspace{2pt}
\resizebox{\linewidth}{12mm}{
\begin{tabular}{cccccc}
\toprule
\toprule
{ Model Variants} &  {O-Precision}      & {O-Recall}         & {B-Precision}      & {B-Recall}         \\
\midrule
{QwQ-32B}       &  { 72.49\%$\pm$0.89\%}          & { 62.30\%$\pm$0.97\%}          & { 57.78\%$\pm$0.65\%}          & { 29.88\%$\pm$0.41\%}          \\
{ w/o Neighborhood-Conditioned Skill Induction}     &  { {\underline{72.54\%$\pm$0.90\%}}}    & { 71.72\%$\pm$0.86\%}          & { 54.00\%$\pm$0.71\%}          & { 31.03\%$\pm$0.43\%}          \\
{ w/o Execution-Aligned Skill Refinement}     &  { 72.36\%$\pm$0.89\%}          & { 71.09\%$\pm$0.85\%}          & { 55.56\%$\pm$0.62\%}          & { 28.73\%$\pm$0.59\%}          \\
{ w/o Context-Budgeted Skill Routing}     &  { 69.83\%$\pm$0.91\%}          & { \textbf{88.02\%$\pm$0.96\%}} & { {\underline{63.39\%$\pm$0.77\%}}}    & { {\underline{40.80\%$\pm$0.55\%}}}    \\
{KDASO\textsubscript{QwQ-32B}}       &  { \textbf{73.42\%$\pm$0.90\%}} & { {\underline{ 87.38\%$\pm$0.97\%}}}    & { \textbf{69.17\%$\pm$0.89\%}} & { \textbf{47.70\%$\pm$0.60\%}}\\
\bottomrule
\bottomrule
\end{tabular}}
\label{rq5}
\vspace{-4pt}
\end{table}

\subsection{Stability of Generated Skills by KDASO}
\label{5.4}
In this section, we investigate the stability of generated skills throughout continuous optimization. To simulate the dynamic progression of real-world scenarios, we adopt a rolling update mechanism, periodically (every one month) generating new skills by KDASO using collected data up to the latest month date and skills generated last month. For example, the August dataset is constructed by combining adjudication data in August. KDASO then generates the August skills based on the August dataset and the July skills. Finally, we evaluate the August skills on the September dataset and then generate the September skills by KDASO on the September dataset and the August skills. By continuously generating skills from July 2025 to December 2025, KDASO demonstrates a significant month-over-month improvement in Figure~\ref{rq3}. This finding indicates that KDASO possesses strong online stability, effectively assimilating real-world adjudication data whose underlying distribution evolves over time, and dynamically refining its adjudication skills accordingly. Consequently, KDASO supports not only high-performance adjudication in static settings but also enables self-evolution of skills and sustained performance gains in dynamic scenarios.

\subsection{Ablation Study}
We conduct an ablation study to quantify the contribution of each module of KDASO in Table~\ref{rq5}. First, removing Neighborhood-Conditioned Skill Induction causes a substantial drop in performance, particularly in B-Precision, which confirms that static skill sets suffer from capability stagnation and fail to capture emerging behavioral patterns in dynamic scenarios. Second, the exclusion of Execution-Aligned Skill Refinement leads to a marked decline in recall, as this module is responsible for analyzing discrepancies between agent execution trajectories and expert logic to transform abstract instructions into more executable, exemplar-based descriptions. Finally, disabling Context-Budgeted Skill Routing results in significant performance decay due to skill overload, demonstrating that in high-skill-density environments, a precise routing mechanism is essential to mitigate the risks of skill omission and uneven attention allocation inherent in long-context processing.

\section{Conclusion}
\label{conclusion}
In this paper, we address three critical limitations in liability adjudication tasks using LLMs: stale skill libraries, semantic execution gaps between skill descriptions and intended expert logic, and context budget constraints leading to critical skill omission. To overcome these challenges, we propose KDASO, which continuously evolves a skill library through neighborhood-conditioned induction, execution-aligned refinement, and context-budgeted routing. Experiments demonstrate that KDASO achieves excellent performance across three real-world liability adjudication tasks and and that the generated skills exhibit high transferability and stability.


Several limitations warrant future investigation. First, the current skill induction operates on local neighborhood structures in embedding space, which may overlook boundary cases that are individually distant but jurisprudentially critical. Second, while we demonstrate promising transferability across conceptually similar scenarios, the framework's robustness under substantial domain shifts, such as cross-regulatory or cross-lingual adjudication, has yet to be validated. 

\section*{Acknowledgements}
This work is supported by  the NSFC (Grant No. 62306133), the Key Program of the Science Foundation (Grant No. BK20243012) and the CCF-DiDi GAIA Collaborative Research Funds for Young Scholars. We also appreciate the guidance provided by the engineers in DiDichuxing Co. Ltd.

\bibliography{main}
\bibliographystyle{plain}


\appendix
\newpage

\addcontentsline{toc}{section}{附录目录} 

\vspace{1em}
\section*{APPENDIX CONTENTS}
\startcontents[appendices]
\printcontents[appendices]{l}{1}{\setcounter{tocdepth}{2}}

\appendix
\onecolumn

\section{Details in KDASO}
\label{KDASO}

For the liability adjudication task $f: (x, \mathcal{S}) \rightarrow y$, we employ the following prompt, using ride-hailing adjudication as an example, which integrates the input case $x$ and skills $\mathcal{S}$ to simultaneously generate the reasoning trajectory $y_T$ and the liability $y_L$.


\begin{minipage}{\columnwidth}
\begin{tcolorbox}[colback=white, colframe=black, title=Prompt - Liability Adjudication, fonttitle=\bfseries, rounded corners]

\# Task Description  

You are a liability attribution expert in a ride-hailing marketplace, responsible for determining the driver’s responsibility in cases where an order is canceled.

\vspace{\baselineskip}

\# Liability Attribution Procedure 

**Step 1. Order Information Extraction**: Order information comprises four primary categories: (i) basic metadata such as origin and destination, estimated pickup time, and fee; (ii) driver-submitted textual reports; (iii) in-app communication records between driver and passenger.  

**Step 2. Liability Scenario Identification**: Based on the extracted order information and in strict adherence to the “Liability Attribution Guidelines,” sequentially evaluate whether the case matches any predefined liability scenarios. Evaluation must follow the prescribed order without omission.  

**Step 3. Integrated Judgment**: Synthesize all available evidence, including order details, driver and passenger behaviors, and verification outcomes, to reach a final liability determination.

\vspace{\baselineskip}

\# Liability Attribution Guidelines  

\vspace{\baselineskip}

\#\# Liability Attribution General skills  

1. **Applicability**: General skills govern the application of all specific verification skills. Prior to scenario-specific verification, first confirm whether the driver submitted a report, and whether in-app calls or messages reference relevant circumstances. Additionally, authenticate the validity of any image evidence.  

2. **Driver-Mention Requirement**: For scenarios involving circumstances such as no-stopping zones, vehicle breakdowns or accidents, government enforcement actions, impassable roads, driver health issues, or electric vehicle power depletion, verification is only triggered if explicitly mentioned by the driver in their report, call, or message. If unmentioned, skip verification for that scenario. For all other scenarios not subject to this mention requirement, each must be systematically evaluated against the verification criteria.  

\vspace{\baselineskip}

\#\#  Liability Attribution Domain-Specific Skills

\textbf{<Skills>}

\vspace{\baselineskip}

\# Output Requirements 

- **Liability Outcome**: The final judgment must be output using the exact XML tag format `<result></result>`.  

  - Example: `<result>Driver Fault</result>`  

- **Liability Scenario**: The matched scenario must be output using the exact XML tag format `<judge></judge>`. The content **must verbatim reproduce the scenario description as listed in the “Liability Attribution Guidelines”** (excluding any parenthetical references), **without paraphrasing, synonym substitution, or modification**, and **must align precisely with the assigned liability category**.

  - Example: `<judge>Skill: Detour</judge>`  

- **Detailed Analysis**: Provide a clear, step-by-step justification for the judgment, including the verification process, evidentiary basis, and logical reasoning. While referencing factual elements from the case is permitted, the original wording of the liability scenarios from the guidelines must remain unchanged. 

\vspace{\baselineskip}

\# Complete Case Information Is Provided Below:

\textbf{<case>}

\end{tcolorbox}
\end{minipage}

\subsection{Neighborhood-Conditioned Skill Induction}
\label{generator}

We detail the implementation of Neighborhood-Conditioned Skill Induction in Algorithm~\ref{alg:KDASO1}. To construct the local manifold structure, we employ BGE-Large-v1.5~\footnote{https://huggingface.co/collections/BAAI/bge} as the embedding model $\phi$ to compute cosine similarities between instances. This approach addresses the tension between single-case overfitting and full-dataset intractability by inducing skills $s_{\text{new}}^i$ from the top-$k$ nearest neighbors $\mathcal{N}_k(x^i)$, ensuring the input complexity remains well within the LLM context window $L_{\text{ctx}}$ while capturing sufficient local evidence.

To mitigate hallucination, we implement a rigorous post-induction validation mechanism rather than relying solely on prompt constraints. Specifically, each induced skill $s_{\text{new}}^i$ is evaluated against its support set $\mathcal{S}_{\text{supp}}(s_{\text{new}}^i) = \{(x, y) \in \mathcal{D}_{\text{new}} \mid f(x, s_{\text{new}}^i) \rightarrow y\}$. Skills with an empty support set are identified as hallucinations and immediately discarded. This ``Conservative Strategy'' aligns with high-stakes industrial requirements, where avoiding erroneous rules (False Positives) is prioritized over maximizing coverage, thereby preventing the deployment of invalid logic.

To further enhance the quality of the skill set and control its scale, we introduced a two-stage pruning strategy. First, we eliminate skills that cannot be applied to any case. Second, for skills that yield inconsistent judgment results, we pair them and re-input them into LLMs, guiding them to generate new skills that are semantically consistent and structurally integrated. This merging process aims to preserve effective discriminative logic while reducing redundancy. For two skills that pass validation and yield identical judgment results, we continue to use the old skills. These old skills can be regarded as high-quality skills validated through multiple rounds, thus warranting retention.

\begin{algorithm}
\caption{Neighborhood-Conditioned Skill Induction}
\label{alg:KDASO1}
\begin{algorithmic}
\STATE \textbf{Input: } $\mathcal{D}_{\text{new}} = \{(x^i, y^i)\}_{i=1}^{N}$ with $y^i = (y_L^i, y_T^i)$, existing skill library $\mathcal{S}$
\STATE \textbf{Output: } Augmented skill library $\mathcal{S}'$
\STATE $\mathcal{S}' \gets \mathcal{S}$
\FOR{each $(x^i, y^i) \in \mathcal{D}_{\text{new}}$}
    \STATE Compute $\text{sim}(x^i, x^j) = \frac{\langle \phi(x^i), \phi(x^j) \rangle}{\|\phi(x^i)\| \cdot \|\phi(x^j)\|}$ for all $j \neq i$
    \STATE $\mathcal{N}_k(x^i) \gets \text{top-}k \text{ most similar instances } \{(x^j, y^j)\}_{j=1}^{k}$
    \STATE $s_{\text{new}}^i \gets \text{LLM}_{\text{generate}}(\mathcal{N}_k(x^i))$
    \vspace{\baselineskip}
    \STATE \textbf{Validation:}
    \STATE Compute $\mathcal{S}_{\text{supp}}(s_{\text{new}}^i) = \{(x, y) \in \mathcal{D}_{\text{new}} \mid f(x, s_{\text{new}}^i) \rightarrow y\}$
    \IF{$\mathcal{S}_{\text{supp}}(s_{\text{new}}^i) = \emptyset$}
        \STATE \textbf{continue} \quad \text{(hallucinated skill, discarded)}
    \ENDIF
    \FOR{each $s_{\text{old}} \in \mathcal{S}'$}
        \IF{$\exists (x', y') \in \mathcal{D}_{\text{new}} \text{ s.t. both skills trigger and yield conflicting outputs}$}
            \STATE Compare empirical accuracies of $s_{\text{new}}^i$ and $s_{\text{old}}$ on overlapping domain
            \IF{$s_{\text{old}}$ has higher accuracy}
                \STATE discard $s_{\text{new}}^i$; \textbf{break}
            \ELSIF{$s_{\text{new}}^i$ has higher accuracy}
                \STATE discard $s_{\text{old}}$; remove $s_{\text{old}}$ from $\mathcal{S}'$; \textbf{break}
            \ELSE
                \STATE $s_{\text{new}}^i \gets \text{LLM}_{\text{merge}}(\{s_{\text{new}}^i, s_{\text{old}}, (x', y')\})$
                \STATE remove $s_{\text{old}}$ from $\mathcal{S}'$
            \ENDIF
        \ENDIF
    \ENDFOR
    \STATE $\mathcal{S}' \gets \mathcal{S}' \cup \{s_{\text{new}}^i\}$
\ENDFOR
\RETURN $\mathcal{S}'$
\end{algorithmic}
\end{algorithm}
\begin{minipage}{\columnwidth}
\begin{tcolorbox}[colback=white, colframe=black, title=Prompt - Generate New Skill, fonttitle=\bfseries, rounded corners]

Please strictly adhere to the following instructions to complete the task. This is a **format-constrained reasoning** task, and any deviation from the specified output format will result in system parsing failure.

\vspace{\baselineskip}

\# Task Objective  

Based on the current liability adjudication case and four historically similar cases, generate two structured textual segments:  

1. An inductive analysis of the adjudication logic employed in the historical cases;  

2. Optimization recommendations for the verification methodology applicable to the current case.

\vspace{\baselineskip}

\# Mandatory Output Specification (Violation of any clause below renders the output invalid)  

1. **Only the following two XML tags are permitted**, and they must appear as matched pairs without nesting:  
   - `<analysis> ... </analysis>`  
   - `<skill> ... </skill>`  
   
2. **Only plain text is permitted within the tags**; inline attributes (e.g., `<analysis id="1">`) and self-closing tags (e.g., `<br/>`) are prohibited.

\vspace{\baselineskip}
\# Content Requirements  

\#\# `<analysis>` Section  

- Focus on analyzing how the “true liability outcome” is derived through factual-chain reasoning, starting from the “order content” and applying the skills specified in the “verification methodology.”  

- Must explicitly incorporate three logical components: spatiotemporal consistency validation, cross-evidence corroboration, and mapping to specific skill provisions.

\#\# `<skill>` Section  

- Output **only** the revised or newly formulated **verification methodology text**.  

- Format: Begin with “ If ”, followed by a precise and actionable description of verification steps.  

- Do not include justifications, references to cases, or meta-commentary (e.g., “It is recommended to revise as…”).

\vspace{\baselineskip}

\# Output Example

<analysis>In similar cases where drivers reported “passenger tardiness,” the platform verified liability by comparing the GPS arrival timestamp with the order creation time to confirm a waiting duration of $\geq$ 5 minutes; audio recordings from the vehicle showed no passenger communication, ruling out driver misjudgment; liability was ultimately assigned to the passenger per Clause 3.2 of the *passenger-Driver Fulfillment Guidelines*.</analysis>  

<skill>If the wait is $\geq$ 5 minutes and no valid communication evidence (e.g., call recordings or in-app messages) exists, the incident is classified as passenger tardiness, output Passenger Liable</skill>

\vspace{\baselineskip}

\# Similar Cases:

\textbf{<Neighbourhood Cases>}

\end{tcolorbox}
\end{minipage}

\begin{minipage}{\columnwidth}
\begin{tcolorbox}[colback=white, colframe=black, title=Prompt - Merge Skill, fonttitle=\bfseries, rounded corners]

Please strictly adhere to the following instructions to complete the task. This is a **format-constrained reasoning** task, and any deviation from the specified output format will result in system parsing failure.

\vspace{\baselineskip}

\# Task Objective  

Merge the provided old and new skills based on the given case to produce an optimized skill formulation such that. The merging process must preserve interpretability and operational feasibility while resolving potential conflicts or redundancies between the original skills.

\vspace{\baselineskip}

\# Mandatory Output Constraints (Violation of any item will render the output invalid)  

1. **Only the following two XML tags are permitted**, and they must appear as matched pairs without nesting:  
   - `<analysis> ... </analysis>`  
   - `<skill> ... </skill>`  
   
2. **Only plain text is permitted within the tags**; inline attributes (e.g., `<analysis id="1">`) and self-closing tags (e.g., `<br/>`) are prohibited.

\vspace{\baselineskip}

\# Content Requirements  

\#\# `<analysis>` Section  

- Focus on analyzing how the semantic and structural characteristics of the input case inform the alignment, conflict resolution, and logical integration of the old and new skills;  

- Must explicitly address: (i) the coverage relationship among skill premises (e.g., inclusion, intersection, mutual exclusion), (ii) the triggering path within the merged skill.

\#\# `<skill>` Section  

- Output **only** the revised or newly formulated **verification methodology text**.  

- Format: Begin with “ If ”, followed by a precise and actionable description of verification steps.  

- Do not include justifications, references to cases, or meta-commentary (e.g., “It is recommended to revise as…”).

\vspace{\baselineskip}

\# Output Example

<analysis>
Old skill emphasizes temporal verification, requiring a minimum waiting duration of $\geq$ 5 minutes based on GPS and order timestamps. New skill claims that drivers falsely claimed passenger tardiness despite active in-app communication. The original skill lacked explicit handling of communication evidence, leading to false positives when drivers ignored incoming messages. The new skill, induced from cases with verified call logs and message histories, introduced the necessity of corroborating the absence of valid communication. By integrating these two dimensions, temporal sufficiency and communicative silence, the merged skill ensures both factual grounding and robustness against strategic misreporting, thereby aligning more closely with expert adjudication practice.
</analysis>

<skill>If the wait is $\geq$ 5 minutes and no valid communication evidence (e.g., call recordings or in-app messages) exists, the incident is classified as passenger tardiness, output Passenger Liable</skill>

\vspace{\baselineskip}

\# Input:

\textbf{<Case>}

\textbf{<Old skill>}

\textbf{<New skill>}

\end{tcolorbox}
\end{minipage}

\newpage

\subsection{Execution-Aligned Skill Refinement}
\label{refiner}
We provide the specific prompt design employed during skill refinement in Algorithm~\ref{alg:KDASO2}. 
To suppress over-hallucination by the LLM in the refinement process, the prompt requires the model to 
output a complete, traceable reasoning chain that explicitly relies on observable features from 
$(x_{s^i}, y_{s^i}, s^i, \mathcal{M}_i)$ and the existing skill logic, rather than introducing 
external or speculative information. 

Additionally, we introduce a semantic similarity-based validation mechanism within refinement. 
For each candidate revised skill $s^{i'}$ generated from $(x_{s^i}, y_{s^i}, s^i, \mathcal{M}_i)$, 
we evaluate its structural and semantic alignment with the original skill $s^i$ and with the 
neighborhood cases used during induction. Let $\text{sim}(\phi(\cdot), \phi(\cdot))$ denote the 
cosine similarity defined in the embedding space $\phi: \mathcal{X} \rightarrow \mathbb{R}^d$. 
If $\text{sim}(\phi(s^{i'}), \phi(s^i)) < \tau_{\text{sem}}$ (a pre‑set semantic threshold), 
the refined skill is flagged as a potential hallucination and is either filtered out or reverted 
to $s^i$. This mechanism constrains the reasonable boundaries of generated results. 

Notably, cosine similarity is employed solely for preliminary screening of execution trajectories 
to determine whether they deviate from the human expert's trajectory ,  specifically, 
filtering out those that are ``similar but may be erroneous'' or ``entirely unrelated''. 
The actual logical refinement is performed by the LLM within the prompt, 
following the iterative process described in \textbf{Execution-Aligned Skill Refinement}, 
until all $G$ trajectories satisfy $c_i^j \geq \tau$ or the maximum iteration budget is reached, 
yielding the refined library $\mathcal{S}''$.

\begin{algorithm}
\caption{Execution-Aligned Skill Refinement in KDASO}
\label{alg:KDASO2}
\begin{algorithmic}
\STATE $\mathcal{S}'' \gets \mathcal{S}'$
\FOR{each $s^i \in \mathcal{S}'$}
    \STATE Find anchor case $(x_{s^i}, y_{s^i}) \in \mathcal{D}_{\text{new}}$ s.t. $f(x_{s^i}, s^i) \rightarrow y_{s^i}$
    \STATE $iter \gets 0$
    \WHILE{$iter < \text{max\_iter}$}
        \STATE Sample $G$ stochastic trajectories $\{y_{s^i}^j\}_{j=1}^{G} \sim \mathbb{P}_{\theta}(\cdot \mid x_{s^i}, s^i)$
        \STATE $\mathcal{M}_i \gets \emptyset$
        \FOR{$j = 1$ \textbf{to} $G$}
            \STATE $c_i^j \gets \text{sim}(\phi(y_{s^i T}^j), \phi(y_{s^i T}))$ \quad (cosine similarity for preliminary screening)
            \IF{$c_i^j < \tau$}
                \STATE $\mathcal{M}_i \gets \mathcal{M}_i \cup \{(y_{s^i T}^j, c_i^j)\}$
            \ENDIF
        \ENDFOR
        \IF{$\mathcal{M}_i = \emptyset$}
            \STATE \textbf{break} \quad (all trajectories aligned)
        \ENDIF
        \STATE $s^{i'} \gets \text{LLM}_{\text{refine}}(x_{s^i}, y_{s^i}, s^i, \mathcal{M}_i)$ \quad (with traceable reasoning chain)
        \STATE \textbf{Semantic validation:}
        \IF{$\text{sim}(\phi(s^{i'}), \phi(s^i)) < \tau_{\text{sem}}$}
            \STATE \textbf{break} \quad (hallucination detected, revert to $s^i$)
        \ENDIF
        \STATE $s^i \gets s^{i'}$
        \STATE $iter \gets iter + 1$
    \ENDWHILE
    \STATE Replace $s^i$ in $\mathcal{S}''$ with final refined skill
\ENDFOR
\RETURN $\mathcal{S}''$
\end{algorithmic}
\end{algorithm}
\newpage

\begin{minipage}{\columnwidth}
\begin{tcolorbox}[colback=white, colframe=black, title=Prompt - Refine Skill, fonttitle=\bfseries, rounded corners] 

Please analyze the input and distill **generalizable experiential knowledge** that can improve future problem-solving performance. Please follow the three-step process below rigorously:

\vspace{\baselineskip}

\# **Step 1: Trajectory Analysis**  

Conduct a thorough review of all provided solution trajectories in light of the ground-truth answer. Specifically:  

- **For successful trajectories**: Identify the key correct decisions, critical insights, and effective reasoning patterns that led to the right solution.  

- **For failed trajectories**: Diagnose precisely where and why the reasoning deviated, e.g., misinterpretation of constraints, overlooked evidence, or flawed inference.  

- **Assess skill gaps**: Determine whether failures stem from (a) relevant skills being absent, (b) existing skills being too vague or misaligned, or (c) skills not being applied at the right decision point.

\vspace{\baselineskip}

\# **Step 2: Skill Update Strategy**

Based on your analysis, refine the current set of skills to better support robust and generalizable reasoning. You need to transform the skills into a logical chain that guides subsequent reasoning. You may perform up to **{max number of operations}** updates to refine an existing skill to enhance clarity, scope, or applicability. 

\#\# For **each proposed update**, you must specify:  

1. **Relevance context**: When this skill would be most useful.  

2. **Key triggering features**: What observable characteristics of a problem signal that this skill applies.  

3. **Generalizability scope**: Other problem types or patterns where this guidance would be transferable.

\vspace{\baselineskip}

\# **Step 3: Skill Formulation Requirements**  

Every skill that is **added or modified** must adhere to the following criteria:  

- Begin with a concise, high-level framing.  

- Focus on **strategic reasoning patterns**, not low-level computations or problem-specific details.  

- Highlight **critical decision points** that commonly arise across similar problems and where guidance can prevent errors or accelerate convergence.

After completing the above reasoning, output your final updates **exclusively** in the following JSON format. Only include the skills you actually modify, add, or delete. It is acceptable to return an empty list if no changes are warranted.

\#\# **Note**: You are not required to use all three operation types. You may choose only one type, or even make no changes at all if the existing skills are sufficient and accurate.

\vspace{\baselineskip}

\# Output Requirements 

- **Refined Skill Name**: The refined skill name must be output using the exact XML tag format `<skill></skill>`.  
  - Example: `<skill>Passenger Tardiness</skill>` 

- **Refined Skill**: The refined skill process must be output using the exact XML tag format `<process></process>`.  
  - Example: 
  
    <process>
    
    1. Check whether the driver arrived at pickup location and waited for a duration of at least 5 minutes. If yes, proceed to Step 2; otherwise, exit.
    
    2. Examine whether any valid communication evidence exists, such as in-app messages or call/audio recordings, indicating the passenger attempted to contact the driver or provided justification for delay. If no such evidence is found, output Passenger Liable; otherwise, exit.
    
    </process>

- **Detailed Analysis**: Provide a clear, step-by-step justification for the refinement, including the verification process, difference between reasoning trajectories in LLMs and human experts. 

\vspace{\baselineskip}

\# Input: \textbf{<case>}
  \textbf{<skill>}
  \textbf{<trajectories>}

\end{tcolorbox}
\end{minipage}

\subsection{Context-Budgeted Skill Routing}
\label{selector}

\vspace{0.2cm}
\noindent \textbf{Feature Encoding and Classifier Training.}
We employ BGE-Large-v1.5 as the embedding model $\phi: \mathcal{X} \rightarrow \mathbb{R}^d$ to encode each case $x^j$ into dense embeddings $\phi(x^j)$, which serve as the feature representation for the XGBoost binary classifiers in Skill Retriever. For each skill $s^i \in \mathcal{S}''$, recall the skill-specific dataset:
\[
\mathcal{D}_{s^i} = \{(x^j, \hat{y}_i^j)\}_{j=1}^{N}, \quad \hat{y}_i^j = \mathbb{I}[f(x^j, s^i) \rightarrow y^j] \in \{0,1\}.
\]
To address potential class imbalance in $\mathcal{D}_{s^i}$, we perform undersampling by randomly selecting a subset of negative samples equal in size to the number of positive samples. Let $N_+ = |\{(x^j, \hat{y}_i^j) \in \mathcal{D}_{s^i} \mid \hat{y}_i^j = 1\}|$ and $N_- = |\{(x^j, \hat{y}_i^j) \in \mathcal{D}_{s^i} \mid \hat{y}_i^j = 0\}|$. The balanced dataset $\tilde{\mathcal{D}}_{s^i}$ is constructed as:
\[
\tilde{\mathcal{D}}_{s^i} = 
\begin{cases}
\{(x^j, \hat{y}_i^j) \mid \hat{y}_i^j = 1\} \cup \text{RandomSample}(\{(x^j, \hat{y}_i^j) \mid \hat{y}_i^j = 0\}, N_+), & \text{if } N_+ < N_-, \\[4pt]
\mathcal{D}_{s^i}, & \text{if } N_+ \geq N_-.
\end{cases}
\]
Based on this balanced training set, the recall of classifier $m^i: \mathcal{X} \rightarrow \{0,1\}$ can exceed $90\%$, meeting real-world liability adjudication scenario requirements. While adjustable in practice, these values are fixed throughout our experiments for consistency.

\vspace{0.2cm}
\noindent \textbf{Retrieval and Structural Matching.}
At test time, given $x_{\text{test}}$, the retrieval stage collects skills predicted as applicable:
\[
\mathcal{S}_{\text{retrieve}} = \{s^i \in \mathcal{S}'' \mid m^i(x_{\text{test}}) = 1\}.
\]
To mitigate classifier false negatives, we augment with direct structural matches:
\[
\mathcal{S}_{\text{match}} = \{s^i \in \mathcal{S}'' \mid x_{\text{test}} \models s^i\},
\]
where $x_{\text{test}} \models s^i$ denotes that $x_{\text{test}}$ satisfies the structural conditions of skill $s^i$. The initial candidate set is:
\[
\mathcal{S}_{\text{final}} = \mathcal{S}_{\text{retrieve}} \cup \mathcal{S}_{\text{match}}.
\]

\vspace{0.2cm}
\noindent \textbf{Context Budget Constraint.}
To respect the context window limit $L_{\text{ctx}}$, we compute the maximum number of storable skills $k_{\max}$ by solving:
\[
k_{\max} = \max \left\{ K \in \mathbb{N} \;\middle|\; \ell(x_{\text{test}}) + \sum_{j=1}^{K} \ell(s^{\sigma(j)}) \leq L_{\text{ctx}} \right\},
\]
where $\sigma: \{1,\dots,|\mathcal{S}_{\text{final}}|\} \rightarrow \mathcal{S}_{\text{final}}$ is a priority ordering that combines classifier confidence $p_i = m^i(x_{\text{test}})$ (the predicted probability of applicability) and direct match signals, such that skills with $x_{\text{test}} \models s^i$ receive higher priority. If $|\mathcal{S}_{\text{final}}| > k_{\max}$, we truncate to the top-$k_{\max}$ skills:
\[
\mathcal{S}_{\text{final}}^{\text{truncate}} = \{\sigma(1), \sigma(2), \dots, \sigma(k_{\max})\}.
\]
The final adjudication is:
\[
y_{\text{test}} = f(x_{\text{test}}, \mathcal{S}_{\text{final}}^{\text{truncate}}),
\]
ensuring completeness and feasibility within the $L_{\text{ctx}}$ budget.

\vspace{0.2cm}
\noindent \textbf{Design Motivation and Scalability.}
This design is motivated by the high cost of false positives in liability adjudication tasks. misapplying an inapplicable skill can lead to erroneous liability attribution with significant operational and reputational consequences. In contrast to dense retrieval methods, which rely on semantic similarity $\text{sim}(\phi(x_{\text{test}}), \phi(x^j))$ and may retrieve skills that are topically related but logically irrelevant, our XGBoost-based approach performs a strict applicability decision based on learned discriminative patterns. In dynamically changing scenarios, corresponding matching skills may lack semantic similarity or appear mismatched in expression. To prevent skill omissions, XGBoost aligns cases with skills at the embedding level via the trained classifiers $m^i$.

Moreover, in liability adjudication scenarios such as ride-hailing adjudication, the total number of domain-specific skills $|\mathcal{S}''|$ typically ranges from $10$ to $100$, smaller than general-purpose document corpora. At this scale, training $|\mathcal{S}''|$ lightweight XGBoost classifiers (one per skill) is computationally feasible and highly parallelizable, requiring $\mathcal{O}(|\mathcal{S}''| \cdot N \cdot d)$ time, enabling efficient real-time skill retrieval without sacrificing accuracy or interpretability. By leveraging this architecture, Skill Retriever significantly reduces the cardinality of $\mathcal{S}''$ to $|\mathcal{S}_{\text{final}}^{\text{truncate}}| \leq k_{\max} \ll |\mathcal{S}''|$, thereby substantially alleviating model reasoning load and mitigating the risk of critical skill omission. For detailed experiments comparing the number of skills, see Appendix~\ref{exp}.

\subsection{Data Usage and Manual Review Mechanism}
Neighborhood-Conditioned Skill Induction and Execution-Aligned Skill Refinement in KDASO share the same recent adjudication dataset $\mathcal{D}_{\text{new}}$. Although this design theoretically entails an overfitting risk, acquiring large-scale, high-quality annotated data is prohibitively expensive in practical liability adjudication tasks, as it typically relies on domain experts for meticulous labeling. Consequently, the system must possess the capability to efficiently generate generalizable skills under limited supervisory signals, and the effective performance of KDASO demonstrates its ability. To mitigate potential overfitting, Neighborhood-Conditioned Skill Induction does not produce skills based on a single sample. Instead, it constructs a neighborhood sample set $\mathcal{N}_k(x)$ via embedding similarity metrics and uses this set as contextual input to LLMs, thereby inducing LLMs to infer more generalizable structured skills. This strategy effectively avoids one-to-one mapping–style skill generation and enhances the applicability of skills to unseen cases.

Regarding the human review mechanism for generated skills, KDASO employs a fully automated pipeline during offline experimental evaluation. Generated skills are directly applied to model reasoning without human intervention, to objectively measure the end-to-end performance of the framework. However, in real-world liability adjudication scenarios, all skills generated by KDASO and intended for incorporation into the knowledge base undergo rigorous manual review by the governance team. This review process includes not only assessing the logical correctness and liable-adjudicated domain-specific alignment of the skills but also performing post-processing operations such as skill merging and pruning. Specifically, in high-stakes adjudication scenarios, the platform commonly adopts a Conservative Strategy prioritizing adjudication precision over recall, tolerating false negatives while strictly avoiding false positives. Therefore, during manual review, a substantial number of skills that output “liable” are proactively filtered out, especially those lacking sufficient evidentiary support or exhibiting ambiguous boundaries, ensuring that the final skill set adheres to the platform’s core requirements for fairness and reliability. More information about Ethics Review can be found in Appendix~\ref{ethic}.

\begin{table}[t]
\centering
\caption{Datasets in DiDi ride-hailing scenarios. All tasks are multi-class classification. Open-source datasets are shown in~\ref{exp:datasets}. Length(tokens) denotes the total sequence length when the text obtained by converting all modalities into texts and concatenating them is fed into the locally deployed tokenizer, excluding skills.}
\vspace{0.1in}
\begin{tabular}{c|cccc}
\toprule
\toprule
Dataset & Samples & Time Span & Length(tokens)\\
\midrule
CA& 1,148 & 2025/10/16 - 2025/10/22  & 1,001\\
CR& 6,200   & 2025/10/01 - 2025/10/31 & 996\\
CPC& 2,455 & 2025/10/09 - 2025/10/21 & 1,914\\
CDC & 2,529 & 2025/10/09 - 2025/10/21 & 1,683\\
FD& 1,358 & 2025/10/01 - 2025/10/27 & 300\\ 
\bottomrule
\bottomrule
\end{tabular}
\label{datasets}
\end{table}

\begin{table}
\centering
\caption{Law-specific and logic-specific datasets. These datasets only contain texts.}
\vspace{0.1in}
\begin{tabular}{cccc}
\toprule
\toprule
Benchmark & Dataset & Samples & Length(tokens)\\
\midrule
\multirow{2}{*}{LawBench} & LA-ITI & 500 & 49.7\\
         & LA-CP  & 500 & 660.6\\
\midrule
\multirow{3}{*}{LegalBench} & LE-OPP & 4,271 & 43.2\\
           & LE-UN  & 3,813 & 44.6\\
           & LE-LH  & 4,992 & 325.3\\
\midrule
Reclor &        & 1,000 & 95.68\\
\midrule
AR-LSAT &       & 230 & 153.9\\
\bottomrule
\bottomrule
\end{tabular}
\label{tab:dataset_stats}
\end{table}

\section{Datasets}
\label{exp:datasets}
This section provides background information and the sources of each dataset evaluated on KDASO and construction methods on LegalBench. The primary attributes of ride-hailing datasets are presented in Table~\ref{datasets} and attributes of law-specific and logic-specific datasets are presented in Table~\ref{tab:dataset_stats}.

\subsection{Ride-hailing Datasets in DiDi}
The Cancel-related scenario refers to a liability determination context triggered by the unilateral cancellation of an order by either the passenger or the driver after a ride-hailing service reservation has been successfully confirmed. This scenario involves behaviors of multiple parties: passengers may cancel orders due to changes in their own itineraries, discovery of better travel options, or other such factors. Drivers may cancel due to sudden vehicle malfunctions, receipt of higher-revenue orders, or other reasons. Liability determination requires comprehensive consideration of numerous factors. Cancellation timing is an important dimension, platforms usually set a free cancellation time window, and cancellations beyond this period may entail liability. The reasonableness of the cancellation reason is also crucial. Reasonable causes such as sudden illness or force majeure may mitigate or exempt liability, whereas unjustified cancellations require bearing corresponding consequences. Meanwhile, the notification method and timeliness cannot be overlooked. Failure to timely and effectively notify the other party of order cancellation and thereby causing losses entails corresponding liability. Regarding adjudication outcomes, passengers who cancel without justification or cancel outside the free cancellation period may need to pay a cancellation fee to compensate drivers for losses. Drivers who cancel without justification or fail to notify passengers as required may face platform penalties and compensate passengers accordingly.

The Fee-related scenario focuses on liability determination issues concerning fee calculation, payment, and disputes during the ride-hailing service process. In terms of fee types, normal fees are calculated according to the platform’s established pricing standards based on factors such as trip distance, travel time, and vehicle type. Additional fees may arise under special circumstances, such as nighttime service, highway tolls, or parking fees~\cite{wu2026progressive}. In the liability determination process, the accuracy of fee calculation must be carefully considered. The platform should ensure that its billing system operates correctly, and if abnormal charges result from system failures or calculation errors, the platform is responsible for making adjustments. The duty to inform about fees is also a critical factor. Drivers are obligated to clearly inform passengers of any potential additional fees before or during the trip, and drivers who fail to fulfill this duty and thereby trigger disputes must bear corresponding liability. Passenger payment behavior is equally important. Passengers should pay the fee promptly as stipulated by the platform, and unjustified refusal or delayed payment entails breach-of-contract liability. Regarding adjudication outcomes, if the platform causes fee-related issues due to its own faults, it must adjust the fee and compensate for losses. Drivers who charge in violation of regulations or fail to fulfill their notification obligations must refund overcharged amounts or provide compensation. Passengers who unjustifiably refuse to pay may face collection actions by the platform and account restrictions.

From the Cancel category, we select four representative and operationally distinct subtasks that reflect common yet nuanced liability adjudication challenges in real-world ride-hailing scenarios:
\begin{itemize}
    \item Cancel–Appeal (CA): a post-cancellation dispute initiated by the driver through a formal appeal, typically contesting an automated or passenger-initiated cancellation ruling and seeking re-evaluation of responsibility.
    \item Cancel–Passenger cancel (CPC): scenarios in which the passenger unilaterally terminates the order after confirmation, raising questions about justification, timing, and compliance with platform policies.
    \item Cancel–Driver cancel (CDC): cases where the driver cancels the confirmed order, often involving assessments of operational necessity, prior notification, and adherence to service obligations.  
    \item Cancel–Reservation (CR): cancellations of pre-scheduled reservation orders that occur before the scheduled pickup time becomes active, which entail distinct policy considerations compared to on-demand rides due to advance commitment and planning.
\end{itemize}

Within the Fee category, we focus on Fee–Detour (FD), a critical subtask concerning fee disputes arising when the actual driving distance significantly exceeds the route distance estimated at booking. Such deviations may stem from traffic conditions, navigation errors, or potentially unjustified detours, necessitating careful evaluation of route legitimacy, driver intent, and routing guidance to determine whether the resulting fee increase is warranted or constitutes overcharging.

Table~\ref{datasets} shows detailed attributes in DiDi ride-hailing scenarios.

\subsection{Law-Specific Datasets}
LawBench has been meticulously crafted to have precise assessment of the LLMs’ legal capabilities. In designing the testing tasks, LawBench simulated three dimensions of judicial cognition and selected 20 tasks to evaluate the abilities of large models. Compared to some existing benchmarks with only multiple-choice questions, we include more diverse types of tasks closely related to real-world applications, such as legal entity recognition, reading comprehension, criminal damages calculation and consultation. LawBench recognizes that the security policies of current large models may decline to respond to certain legal queries or experience difficulty in comprehending instructions, leading to a lack of response. 

\begin{itemize}
    \item Issue Topic Identification (LA-ITI): This task involves categorizing a user’s inquiry into one of several predefined legal topics. Since user queries are often ambiguous or imprecise, correctly mapping them to relevant legal domains enables legal practitioners to more effectively identify the core legal issues at hand. The data for this task is sourced from the CrimeKgAssistant project. To construct the evaluation set, LawBench retains the 20 most frequently occurring topic classes and randomly select 25 questions from each class, resulting in a balanced test set.
    \item Charge Prediction (LA-CP): This task requires predicting the cause of action (i.e., the formal legal charge) based on the factual description extracted from a judicial judgment and the corresponding cited legal article. The cause of action, as defined by the people’s court, encapsulates the essential nature of the legal relationship underlying a litigation case. Accurate prediction of this label supports more efficient and equitable judicial administration. Specifically, during case filing and adjudication, such predictions can assist courts in appropriately assigning cases, allocating judicial resources, and scheduling hearings, thereby enhancing both efficiency and fairness in the legal process. For this task, LawBench draws a sample of 500 instances from the CAIL2018 dataset dedicated to cause-of-action prediction.
\end{itemize}

LegalBench is a collaboratively constructed benchmark comprising 162 fine-grained legal reasoning tasks, developed in close partnership with legal professionals to systematically evaluate LLMs on authentic legal competencies. The dataset integrates three sources: (1) restructured existing legal datasets adapted to few-shot prompting; (2) previously unpublished, expert-crafted datasets from legal scholars; and (3) newly designed tasks targeting diverse legal skills and real-world applications. Tasks span core legal reasoning types, including issue spotting, rule application, interpretation, and rhetorical analysis, and cover scenarios such as contract clause classification, hearsay determination, jurisdictional assessment, legal definition extraction, and securities litigation element identification. All instances are derived from real legal documents (e.g., contracts, court opinions, complaints, privacy policies) or meticulously authored by legal experts to reflect professional standards. With labels encompassing classification, entity extraction, and open-ended generation, LegalBench offers a high-quality, multidimensional evaluation framework that balances professional rigor with practical relevance, enabling comprehensive assessment of LLMs’ legal understanding, reasoning, and applicability.

\begin{itemize}
    \item OPP-115 (LE-OPP): This task evaluates LLMs' ability to comprehend and classify data practices in online privacy policies. The tasks are based on the OPP-115 Corpus, a collection of 115 natural-language privacy policies meticulously annotated by a team of three law graduate students to identify ten distinct categories of data handling practices: First Party Collection/Use, Third Party Sharing/Collection, User Choice/Control, User Access/Edit \& Deletion, Data Retention, Data Security, Policy Change, Do Not Track, International \& Specific Audiences, and Other. For task construction, binary classification subtasks were created for the first nine categories, excluding text segments under 10 words and omitting the overly broad “Other” category to maintain classification specificity. 
    \item Unfair Terms of Service  (LE-UN): This task evaluates LLMs' ability to perform fine-grained multiclass classification of potentially unfair clauses in terms-of-service agreements. Drawing on European consumer law, the task is structured around eight categories of clauses deemed “potentially unfair”: Arbitration, Unilateral Change, Content Removal, Jurisdiction, Choice of Law, Limitation of Liability, Unilateral Termination, and Contract by Using. Given a clause excerpt, the model must determine whether it falls into one of these categories and, if so, assign the correct label.
    \item Learned Hand  (LE-LH): This task constitutes a suite of legal issue-spotting benchmarks, designed to evaluate a model’s capacity to identify latent legal issues from informal, colloquial narratives posted by users on the Reddit subreddit legaladvice. Although these texts eschew formal legal terminology, they often embed multiple, overlapping legal concerns. The tasks are grounded in the Legal Issue Spotting Taxonomy readable, multi-label classification framework derived from legal aid practice and encompassing 16 core legal domains, including housing, employment, and immigration. Annotations were generated through the crowdsourced “Learned Hands” gamified platform, wherein legal professionals performed iterative binary judgments (yes/no/skip) to achieve inter-annotator consensus. The central challenge lies in requiring models to accurately infer applicable legal categories from ambiguous, unstructured natural language, a capability that mirrors the issue-spotting skill exercised by attorneys during initial client intake.
\end{itemize}

We restructured two binary classification tasks from LegalBench: OPP-115 and Learned Hand. Specifically, an analogous procedure was applied to the 9 binary OPP-115, which correspond to data practice categories in privacy policies. Positive instances were relabeled with their category names, multi-label cases were preserved, and the same partial-match scoring rule was employed. For the 16 binary Learned Hand, each corresponding to a distinct legal issue category, we replaced the original Boolean labels (“True”/“False”) with the respective category names. Instances annotated with multiple legal issues were converted into multi-label samples by concatenating all applicable categories. During evaluation, a “partial match” criterion was adopted: a prediction was deemed correct if it matched any one of the ground-truth labels. This reformulation transformed the original ensembles of binary tasks into two unified multi-class (16-way and 9-way) classification datasets.

\subsection{Logic-Specific Datasets}

ReClor is a reading comprehension dataset requiring logical reasoning, which is constructed by collecting high-quality logical reasoning questions from standardized graduate admission examinations such as the GMAT and LSAT. The dataset contains 6,138 multiple-choice questions, each comprising a context, a question, and four answer options, designed to evaluate models’ ability to perform multi-step reasoning, such as identifying implicit conflicts, assumptions, or information that strengthens or weakens an argument. To comprehensively assess the logical reasoning capability of models and mitigate the influence of dataset biases, the test set is partitioned into an EASY subset and a HARD subset: data points that can be correctly answered using only the answer options, indicating potential bias exploitation, are grouped into the EASY set, while the remaining instances constitute the HARD set. Empirical results demonstrate that state-of-the-art models achieve high accuracy on the EASY set but perform near random-chance levels on the HARD set, whereas human subjects maintain consistent performance across both subsets, highlighting the challenge posed by ReClor in advancing genuine logical reasoning in machine reading comprehension.

AR-LSAT is a dataset for analytical reasoning constructed from LSAT logic games. It contains 2,046 multiple-choice questions requiring models to perform constraint satisfaction and deductive reasoning over structured scenarios (ordering, grouping, assignment). Unlike standard reading comprehension, it demands explicit logical deduction rather than semantic matching. Original experiments showed that pre-trained language models struggle significantly (near random performance), highlighting the gap between statistical pattern matching and genuine logical reasoning.

\section{Models}
\label{exp:models}
In this section, we provide introductions to liable-adjudicated general-purpose algorithms and liable-adjudicated domain-specific models. All training experiments are conducted on a cluster of 8 NVIDIA H200 GPUs. All results are reported as the averaged over runs with three random seeds.

\subsection{Liable-Adjudicated General-Purpose Algorithms}
\paragraph{NL2FOL.} 
NL2FOL~\cite{lalwani2025autoformalizing} is a neurosymbolic hybrid approach that automatically formalizes natural language into first-order logic. Its core pipeline consists of three modules. First, through multi-step LLMs prompting, the input sentence is decomposed into premises and conclusions, from which referring expressions and attributes are extracted. Natural language inference techniques are employed to identify relationships among entities and background knowledge among attributes, thereby constructing a structured logical representation that incorporates implicit commonsense. Second, the generated first-order logic formulas are automatically compiled into SMT-LIB formatted SMT files for processing by formal solvers. Finally, an SMT solver is used to verify the validity of the logical formula. If the negation of the formula is satisfiable, it is deemed a logical fallacy, and the counterexample model returned by the solver is translated back into natural language via LLMs to enhance interpretability.

\paragraph{NeoBERT.} 
NeoBERT~\cite{le2025neobert} is a next-generation bidirectional encoder model whose methodological design integrates modern architecture, large-scale data, and an optimized pretraining strategy. Architecturally, it replaces conventional absolute position embeddings with Rotary Position Embedding to support long-context modeling, and introduces RMSNorm with Pre-Layer Normalization to enhance training stability. The SwiGLU activation function is employed to strengthen nonlinear representational capacity. Moreover, by adjusting the depth-to-width ratio, NeoBERT achieves parameter-efficient utilization while maintaining an embedding dimension compatible with BERT-base. In terms of data, NeoBERT is pretrained on the large-scale RefinedWeb web text corpus using a two-stage training strategy: first training on 2 trillion tokens with a sequence length of 1024, then extending to 4096 tokens for an additional 100 billion tokens to efficiently improve contextual processing capability. The pretraining objective retains only masked language modeling with a masking rate of 20 percent, optimized via the AdamW optimizer coupled with cosine learning rate decay scheduling. The overall approach emphasizes plug-and-play compatibility, training sufficiency, and computational efficiency, establishing a robust encoder foundation without reliance on complex fine-tuning techniques.

\paragraph{ModernBERT.} 
ModernBERT~\cite{warner2025smarter} is a modernized bidirectional encoder model whose methodological design integrates architectural enhancements, efficient training strategies, and large-scale data. Architecturally, it adopts a pre-normalization structure, employs the GeGLU activation function, and removes bias terms from all linear layers and LayerNorm modules to optimize parameter allocation. Rotary Position Embedding is used for positional encoding, natively supporting long-context modeling up to 8192 tokens. To enhance efficiency on long sequences, its attention layers follow an alternating pattern where only one out of every three layers uses global attention while the remaining two layers employ local attention with a 128-token sliding window. The model also fully supports unpadding by removing padding tokens once before the embedding layer, enabling padding-free computation through FlashAttention's variable-length attention combined with Rotary Position Embedding, and re-padding the output only when necessary, thereby significantly reducing redundant computation. In training, ModernBERT is pretrained on two trillion tokens of diverse corpora, utilizes an improved BPE tokenizer, and adopts a masked language modeling objective with a 30 percent masking rate. The overall design emphasizes hardware awareness such as tensor core alignment, inference speed, and memory efficiency, while maintaining an embedding dimension compatible with BERT-base to enable seamless transfer.

\paragraph{SIM-RAG.} 
Sim-RAG~\cite{yang2025knowing} is a self-improving RAG framework designed for domain adaptation. Its core methodology consists of two stages. In the first stage, LLMs undergo retrieval-oriented instruction fine-tuning on general-domain data, encompassing four types of tasks: general instruction following, context-aware question answering, answer generation, and query generation, thereby endowing the model with the joint capability to perform question answering and question formulation using external context. In the second stage, the fine-tuned model performs self-training on unlabeled corpora in the target domain. Specifically, it first generates diverse candidate answers from documents, then constructs corresponding questions based on these answers and the source documents, thereby producing a large number of synthetic question-document-answer triplets. High-quality samples are selected through a round-trip consistency filtering strategy. Finally, these pseudo-labeled data are combined with the data from the first stage to further fine-tune the model, achieving adaptive optimization for domain-specific retrieval-augmented generation tasks.

\paragraph{NodeRAG.} 
NodeRAG~\cite{xu2025noderag} is a retrieval-augmented generation framework built upon a fine-grained heterogeneous graph comprising seven node types, entities (N), relations (R), semantic units (S), attributes (A), high-level elements (H), high-level overviews (O), and raw text chunks (T), each serving a distinct role to enable decoupled, hierarchical information representation. The approach proceeds in two phases: graph indexing and graph retrieval. During indexing, LLMs decompose raw text into semantic units, entities, and relations. Key entities are identified via K-core and betweenness centrality and summarized with attributes. Communities are detected using the Leiden algorithm to derive high-level elements and overviews. Raw text chunks are reintegrated, and informative nodes (T, S, A, H) are selectively embedded, with semantic neighbor edges added via HNSW to strengthen connectivity. In retrieval, a dual-path mechanism combines exact matching for entities and overviews with vector similarity search for informative nodes to establish entry points. Shallow personalized PageRank then performs limited random walks from these points to identify query-relevant cross nodes. Only retrievable node types (T, S, A, H, R) are passed as context to LLMs, enabling precise, interpretable, and efficient multi-hop reasoning.

\subsection{Legal-Specific Models}
\paragraph{Lawyer-LLAMA.}
Lawyer LLaMA~\cite{huang2023lawyer} is characterized by a three-stage training paradigm: (1) Domain knowledge injection through continual pre-training on a large-scale Chinese legal corpus, including statutes, judicial interpretations, and court judgments, to embed foundational legal knowledge into the base LLaMA architecture; (2) Professional skill acquisition via supervised fine-tuning on high-quality expert-authored data, such as annotated solutions to National Judicial Examination questions, which empirically demonstrates that hundreds of expert-written examples significantly outperform tens of thousands of ChatGPT-generated ones in teaching nuanced legal reasoning; and (3) Retrieval-augmented generation with irrelevant information filtering, where a legal article retriever provides external evidence during inference, and the model is explicitly trained to discern and disregard irrelevant or misleading retrieved content, thereby mitigating hallucinations and enhancing response reliability. Collectively, this integrated approach enables Lawyer LLaMA to exhibit superior legal knowledge grounding, professional reasoning fidelity, and robustness against spurious information compared to general-purpose or naively adapted LLMs.

\paragraph{LexiLaw.}
LexiLaw~\cite{LexiLaw} is a fine-tuned Chinese LLM specialized for the legal domain, built upon the ChatGLM-6B architecture. It is developed by further training the base model on a curated corpus comprising diverse legal texts, including statutes, judicial interpretations, court judgments, and legal Q\&A pairs. Grounded in deep learning methodologies, LexiLaw is designed to capture both the lexical precision and structural reasoning patterns inherent in legal discourse. As a result, it is capable of generating legally coherent explanations, offering context-sensitive advisory responses, and assisting users in understanding complex legal provisions, thereby serving as a valuable tool for legal information retrieval, preliminary consultation, and educational support within the Chinese legal system.

\paragraph{ChatLaw.}
ChatLaw~\cite{cui2024chatlaw} integrates a knowledge graph–enhanced Mixture-of-Experts architecture with a multi-agent collaborative framework. Its design features three core components: (1) a high-quality, expert-curated Chinese legal dataset spanning 44 fine-grained tasks, structured with knowledge graphs to ensure factual accuracy and comprehensive coverage; (2) an MoE-based language model that dynamically activates specialized “virtual legal experts” to handle diverse legal subtasks, thereby improving response precision; and (3) a standardized operating procedure inspired by real-world law firm workflows, in which four distinct agents, Legal Assistant, Legal Researcher, Senior Lawyer, and Legal Editor, sequentially perform information gathering, legal retrieval, advisory reasoning, and document drafting, effectively mitigating hallucinations and enhancing the completeness, correctness, guidance, and authority of legal consultations.

\subsection{Logic-Specific Models}
\paragraph{MERIt.}
MERIt~\cite{jiao2022merit} is a self-supervised pre-training framework designed to enhance logical reasoning capabilities in language models. Its core innovations include: (1) a meta-path–guided strategy to automatically extract implicit logical structures from unlabeled text and construct semantically consistent positive instance pairs; (2) a relation-editing mechanism that generates negative samples by perturbing logical relations along the meta-path, thereby enforcing the model to discern logical consistency; and (3) a counterfactual data augmentation technique that introduces world-knowledge–violating yet logically coherent examples during pre-training, compelling the model to rely on contextual logical relationships rather than surface-level statistical cues or factual priors, thus mitigating information shortcuts and improving generalization in logical inference.

\paragraph{LReasoner.}
LReasoner Ensemble~\cite{wang2022logic} is a neuro-symbolic logical reasoning framework that integrates explicit symbolic inference with neural language modeling. Its core design comprises two key components: (1) a logic-driven context extension mechanism, which follows a three-stage paradigm, first extracting elementary logical expressions from text, then applying propositional logic laws (e.g., contraposition and transitivity) to derive implicit logical relations, and finally verbalizing these inferred expressions into natural language to augment the original context for downstream matching; and (2) a logic-driven data augmentation strategy, which constructs contrastive training instances by deliberately modifying logical expressions (e.g., through deletion, reversal, or negation) to generate lexically similar but logically inconsistent negative samples, thereby encouraging the model to focus on logical structure, particularly negation and conditional relationships, rather than superficial semantic cues.

\paragraph{IDOL.}
IDOL~\cite{xu2023idol} is a self-supervised pre-training framework designed to enhance logical reasoning capabilities in language models. Its core innovation lies in leveraging six types of logical indicators, including premise/conclusion, negation, adversative, coordinating, and logic-unrelated markers, to construct a large-scale, logically rich pre-training corpus named LGP. During pre-training, the model is tasked with a Logical Category Prediction objective: given a text where indicator tokens are masked with a special token [LGMASK]. This multi-task learning approach, combined with standard Masked Language Modeling, explicitly guides the model to attend to logical structures, such as negation and conditional relationships, without relying on external knowledge graphs or complex symbolic systems, thereby improving its ability to perform robust and generalizable logical inference.

\paragraph{AMR-LDA.}
AMR-LDA~\cite{bao2024abstract} comprises three components: (1) parsing natural language sentences into Abstract Meaning Representation (AMR) graphs, which explicitly encode logical structures such as negation and conditionals; (2) applying four classical propositional logic equivalence laws, contraposition, implication, commutativity, and double negation, to symbolically manipulate the AMR graphs and generate logically equivalent or non-equivalent variants; and (3) converting the modified AMR graphs back into fluent natural language. This approach is architecture-agnostic: it can be integrated with discriminative LLMs via contrastive learning on augmented sentence pairs, or with generative LLMs through prompt augmentation, thereby improving their ability to reason over logical relationships without relying on templates or task-specific structures.

\section{Metrics}
\label{exp:metrics}
We strictly select evaluation metrics in accordance with the official requirements of each benchmark. Let \( y_i \in \{0,1\} \) denote the ground-truth label of the \( i \)-th sample and \( \hat{y}_i \in \{0,1\} \) its predicted label, for \( i = 1, \dots, N \), where $N = \text{Number of Tests}$. The indicator function \( \mathbb{I}(\cdot) \) equals 1 if the condition is true and 0 otherwise.

\begin{itemize}
    \item Accuracy: LA-ITI in LawBench, LE-OPP and LE-LH in LegalBench, AR-LSAT.
    \[
      \text{Accuracy} = \frac{1}{N} \sum_{i=1}^{N} \mathbb{I}(y_i = \hat{y}_i).
    \]
    Multi-class (\(K>2\)):
    \[
      \text{Accuracy} = \frac{1}{N} \sum_{i=1}^{N} \mathbb{I}(y_i = \hat{y}_i).
    \]
    
    \item Balanced Accuracy: LE-UN in LawBench.
    \[
      \text{Balanced Accuracy} = \frac{1}{2} \left( 
        \frac{\sum_{i=1}^{N} \mathbb{I}(y_i = 1 \land \hat{y}_i = 1)}{\sum_{i=1}^{N} \mathbb{I}(y_i = 1)} +
        \frac{\sum_{i=1}^{N} \mathbb{I}(y_i = 0 \land \hat{y}_i = 0)}{\sum_{i=1}^{N} \mathbb{I}(y_i = 0)}
      \right),
    \]
    Multi-class (\(K>2\)):
    \[
      \text{Balanced Accuracy} = \frac{1}{K} \sum_{k=1}^{K} \frac{\sum_{i=1}^{N} \mathbb{I}(y_i = k \land \hat{y}_i = k)}{\sum_{i=1}^{N} \mathbb{I}(y_i = k)}.
    \]
    
    \item F1-score: LA-CP in LawBench.
    \[
      \text{F1} = \frac{2 \cdot \left( \sum_{i=1}^{N} \mathbb{I}(y_i = 1 \land \hat{y}_i = 1) \right)}
                       {\sum_{i=1}^{N} \mathbb{I}(y_i = 1) + \sum_{i=1}^{N} \mathbb{I}(\hat{y}_i = 1)},
    \]
    Multi-class (\(K>2\), Macro-F1):
    \[
      \text{Macro-F1} = \frac{1}{K} \sum_{k=1}^{K} \frac{2 \cdot P_k \cdot R_k}{P_k + R_k}, \quad
      P_k = \frac{TP_k}{TP_k + FP_k}, \quad R_k = \frac{TP_k}{TP_k + FN_k},
    \]
    where \(TP_k = \sum_{i=1}^{N} \mathbb{I}(y_i = k \land \hat{y}_i = k)\), \(FP_k = \sum_{i=1}^{N} \mathbb{I}(y_i \neq k \land \hat{y}_i = k)\), \(FN_k = \sum_{i=1}^{N} \mathbb{I}(y_i = k \land \hat{y}_i \neq k)\).
    
    \item Precision and Recall: CA, CR, CPC, CDC, FD in DiDi standards.
    \[
      \text{Precision} = \frac{\sum_{i=1}^N \mathbb{I}(y_i = 1 \land \hat{y}_i = 1)}{\sum_{i=1}^N \mathbb{I}(\hat{y}_i = 1)}, \quad
      \text{Recall} = \frac{\sum_{i=1}^N \mathbb{I}(y_i = 1 \land \hat{y}_i = 1)}{\sum_{i=1}^N \mathbb{I}(y_i = 1)}.
    \]
    Multi-class (\(K>2\), Macro-averaged):
    \[
      \text{Macro-Precision} = \frac{1}{K} \sum_{k=1}^{K} P_k, \quad
      \text{Macro-Recall} = \frac{1}{K} \sum_{k=1}^{K} R_k,
    \]
    where \(P_k\) and \(R_k\) are defined as above.
    
    \item Official online evaluation API: ReClor\footnote{\url{https://eval.ai/web/challenges/challenge-page/503/submission.}}.
\end{itemize}

\begin{table}
\centering
\caption{Details of Skill Changes from August 2025 to December 2025 on KDASO.}
\vspace{0.1in}
\begin{tabular}{cccccc}
\toprule
\toprule
Month & 2025.08 & 2025.09 & 2025.10 & 2025.11 & 2025.12 \\
\midrule
Number of $\mathcal{D}_{new}$ & 3000 & 3000 & 3000 & 3000 & 3000 \\
Number of \textbf{Generated} Skills & 12 & 15 & 8 & 6 & 6 \\
Number of \textbf{Merged} Skills & 4 & 4 & 3 & 1 & 2 \\
Number of \textbf{Dropped New} Skills & 5 & 6 & 5 & 4 & 3 \\
Number of $\mathcal{R}'$ & 78 & 83 & 83 & 84 & 85 \\
Average Length of $\mathcal{R}''$ & 137.4 & 146.3 & 147.5 & 149.1 & 149.8 \\
Average Number of $\mathcal{R}_{final}$ & 12 & 13 & 13 & 12 & 13 \\
\bottomrule
\bottomrule
\label{continue}
\end{tabular}
\end{table}

\section{Details on Continual Skill Optimization}
\label{exp}
Continual fine-tuning analysis refers to a monthly iterative process in which a new set of adjudication cases \(\mathcal{D}_{\text{new}}\) is combined with \(\mathcal{S}\) and fed into KDASO to generate and refine \(\mathcal{S}''\), which is then evaluated on the following month’s cases. We conduct this analysis on CA, starting from an initial skill base of 75 human-authored skills, each 100–200 tokens long.

As shown in Table~\ref{continue}, over the period from August 2025 to December 2025, the number of newly generated skills per month generally decreases, suggesting that the adjudication scenarios evolve slowly and that the skill set gradually stabilizes. Concurrently, the number of merged skills, those consolidated with existing ones via skill Refiner, also declines, indicating that KDASO progressively sharpens skill distinctions, potentially reducing redundancy or overlap. Meanwhile, KDASO consistently discards a modest number of newly generated skills, likely due to low relevance, poor coverage, or inconsistency with observed patterns. This suggests a built-in mechanism for quality control during skill evolution.

Notably, both the total number of skills in \(\mathcal{S}''\) and their average length show a mild but steady upward trend. This implies that KDASO incrementally enriches the skill base, not through uncontrolled expansion, but by adding nuanced refinements and elaborating skill descriptions to better capture emerging case patterns. Moreover, Context-Budgeted Skill Routing don't yield an increased number of retrieved skills as $\mathcal{S}''$ grew. Instead, it consistently retrieved around 12 skills, further validating the effectiveness of Context-Budgeted Skill Routing. Overall, the results suggest that KDASO supports stable, interpretable, and gradual skill evolution under continual learning.


\section{Cost-Effectiveness Analysis}
\label{G}
We compare the computational cost and inference efficiency of LLMs in performing liability adjudication tasks before and after integrating KDASO. As shown in Table~\ref{time}, KDASO substantially reduces the input length, total token consumption, and per-inference runtime of LLMs. This efficiency gain primarily stems from KDASO’s mechanism design. First, Execution-Aligned Skill Refinement semantically refines and logically normalizes the original skills, ensuring that the generated prompts are highly focused and instructionally unambiguous, which effectively suppresses redundant or speculative reasoning by LLMs along irrelevant or ambiguous paths. Second, Context-Budgeted Skill Routing performs precise routing of input cases based on structured adjudication skills, assigning each case only to the most relevant subset of sub-scenario skills, thereby drastically reducing the number of skills fed into LLMs. This demonstrates that KDASO can effectively alleviate the computational burden of LLMs in complex governance tasks, providing a strong foundation for high-throughput, low-latency industrial deployment.

\begin{table}
\centering
\caption{Tokens vs. Runtime.}
\vspace{0.1in}
\label{time}
\begin{tabular}{cccc}
\toprule
\toprule
Model & Input tokens & Output tokens & Time Cost(seconds per inference) \\
\midrule
QwQ-32B & 27,156 & 1,540 & 14.79 \\
KDASO\textsubscript{QwQ-32B} & 7,241 & 1,769 & 12.13 \\
\bottomrule
\bottomrule
\end{tabular}
\end{table}

\section{Hyperparameter Sensitivity Analysis}
\label{H}
This section primarily investigates the impact of key hyperparameters in KDASO, with a focus on the number of neighbors \(k\) in \(\mathcal{N}_k(x^i)\) of Neighborhood-Conditioned Skill Induction and the similarity score threshold \(\tau\) in Execution-Aligned Skill Refinement, which determines whether the generated skills align with the actual adjudication reasoning of human experts. The performance of KDASO on the CA dataset serves as the unified evaluation criterion for assessing quality of all hyperparameter.

\begin{table}[t]
\centering
\caption{Model performance under different $k$ of KDASO\textsubscript{QwQ-32B}.}
\vspace{0.1in}
\label{k}
\begin{tabular}{ccccc}
\toprule
\toprule
$k$ & O-Precision & O-Recall & B-Precision & B-Recall \\
\midrule
QwQ-32B & 72.49\% & 62.30\% & 57.78\% & 29.88\% \\
$k$ = 1 & 65.52\% & 85.62\% & 49.17\% & 33.91\% \\
$k$ = 2 & 65.57\% & 86.10\% & 51.13\% & 39.08\% \\
$k$ = 3 & 66.62\% & 83.55\% & 40.80\% & 29.31\% \\
$k$ = 4 & 72.49\% & 88.82\% & 71.14\% & 60.92\% \\
$k$ = 5 & 64.88\% & 85.30\% & 45.80\% & 34.48\% \\
\bottomrule
\bottomrule
\end{tabular}
\end{table}

\begin{table}[t]
\centering
\caption{Model performance under different $\tau$ of KDASO\textsubscript{QwQ-32B}}
\vspace{0.1in}
\label{tau}
\begin{tabular}{ccccc}
\toprule
\toprule
$\tau$ & O-Precision & O-Recall & B-Precision & B-Recall \\
\midrule
QwQ-32B & 72.49\% & 62.30\% & 57.78\% & 29.88\% \\
$\tau$ = 0.7 & 63.26\% & 88.02\% & 48.73\% & 44.25\% \\
$\tau$ = 0.8 & 62.16\% & 88.18\% & 50.44\% & 32.76\% \\
$\tau$ = 0.9 & 71.67\% & 68.69\% & 58.03\% & 37.36\% \\
\bottomrule
\bottomrule
\end{tabular}
\end{table}

\begin{table}[t]
\begin{center}
\caption{Hyperparameter Grids of XGBoost. }
\label{Hyperparameter_tree_model}
\begin{tabular}{cc}
\toprule
\toprule
Hyperparameter & Values \\ 
\midrule
Learning Rate & $\{0.01, 0.1\}$ \\
Max. Depth & $\{1, 5, 9\}$ \\
N Estimators & $\{10,000, 20,000, 30,000\}$ \\
Subsample & $\{0.5, 0.8, 1.0\}$ \\
Colsample Bytree & $\{0.5, 0.8, 1.0\}$ \\
Min Child Weight & $\{1, 3, 5\}$ \\
\bottomrule
\bottomrule
\end{tabular}
\end{center}
\end{table}

For \(k\) in \(\mathcal{N}_k(x^i)\) of Neighborhood-Conditioned Skill Induction, we conduct an evaluation over the range \(k = 1, 2, 3, 4, 5\). Notably, when \(k\) is increased to 6 or beyond, the length of \(\mathcal{N}_k(x^i)\) exceeds the context window capacity of LLMs due to their inherent context length limitations, resulting in a high failure rate in skill generation. Therefore, we do not include results for \(k \geq 6\) in the table. As shown in Table~\ref{k}, results indicate that \(k = 4\) achieves the best trade-off across all metrics, effectively leveraging semantic information from neighboring cases while avoiding degradation in skill generation quality caused by context overload. Consequently, we fix \(k = 4\) in KDASO.

\label{pr}
We also examine the influence of \(\tau\) in Execution-Aligned Skill Refinement on the effectiveness of skill refinement. \(\tau\) governs the degree of alignment between the logical structure of existing skills and the actual reasoning patterns employed by human experts, thereby modulating the scope of skill reuse and iterative refinement. Notably, to prevent over-optimization, we cap the maximum number of refinement iterations per skill at 20. Table~\ref{tau} compares model performance under different \(\tau\). Based on both quantitative performance and qualitative analysis of the actual generated skills, we observe that when \(\tau\) is set too high, generated skills become overly strict, favoring logical conjunctions (e.g. \(\land\)), which reduces their applicability and makes it difficult to trigger a judgment on liability. Conversely, when \(\tau\) is too low, the skills tend to rely excessively on disjunctive or overly permissive structures (e.g., \( \lor \)), leading the model to prematurely output liable upon satisfying only a few initial conditions. Consequently, we fix \(\tau = 0.8\) in KDASO.

In Context-Budgeted Skill Routing, we adopt XGBoost as the classifier. XGBoost~\cite{10.11452939672.2939785} is an efficient and flexible machine learning model that incrementally builds multiple decision trees by optimizing the loss function, with each tree correcting the errors of the previous one to continuously improve the model's predictive performance. XGBoost also incorporates the gradient boosting algorithm, iteratively training decision tree-based models with the goal of minimizing residuals and enhancing predictive accuracy. The hyperparameter grid is provided in Table~\ref{Hyperparameter_tree_model}.

\begin{table}[t]
\centering
\caption{Changes in metrics by changing $\mathcal{D}_{\text{new}}$ to KDASO\textsubscript{QwQ-32B}. FP and FN are obtained by applying \(\mathcal{R}\) and latest cases on \(f\) to get the confusion matrix.}
\vspace{0.1in}
\label{data_pr}
\begin{tabular}{ccccc}
\toprule
\toprule
$\mathcal{D}_{\text{new}}$ Composition & O-Precision & O-Recall & B-Precision & B-Recall \\
\midrule
FN only & 67.19\%$\pm$0.82\% & 81.47\%$\pm$0.93\% & 47.62\%$\pm$0.60\% & 40.23\%$\pm$0.54\% \\
FP only & 71.70\%$\pm$0.85\% & 72.04\%$\pm$0.86\% & 44.90\%$\pm$0.58\% & 25.29\%$\pm$0.37\% \\
FP+FN & 69.07\%$\pm$0.83\% & 76.36\%$\pm$0.89\% & 43.64\%$\pm$0.57\% & 27.59\%$\pm$0.38\%\\
\bottomrule
\bottomrule
\end{tabular}
\end{table}

\begin{table}[t]
\centering
\caption{Comparison of different retrieval methods.}
\vspace{0.1in}
\label{xgb1}
\begin{tabular}{cccc}
\toprule
\toprule
Method & Precision & Recall & Training Time \\
\midrule
XGBoost & 67.37\% & 89.46\% & 345.56s \\
Dense Retrieval (Bi-Encoder) & 36.10\% & 92.53\% & 0.69s \\
Cross-Encoder & 47.10\% & 94.72\% & 16.34s \\
\bottomrule
\bottomrule
\end{tabular}
\end{table}

\begin{table}[t]
\centering
\caption{Scalability stress test on different numbers of skills.}
\vspace{0.1in}
\label{xgb2}
\begin{tabular}{cccc}
\toprule
\toprule
Skill Count & XGBoost (Parallel) [ms] & Dense Retrieval [ms] & Cross-Encoder (Est.) [ms] \\
\midrule
100 & 7.21 & 5.1 & \\
500 & 36.03 & 5.5 & \\
1,000 & 72.07 & 6.0 & > 1,000 \\
2,000 & 144.14 & 7.0 & \\
5,000 & 360.34 & 10.0 & \\
\bottomrule
\bottomrule
\end{tabular}
\end{table}

To justify the use of cosine similarity for neighborhood construction and trajectory-divergence measurement, we compare four embedding backbones on LA-ITI. We report three metrics: Label Homogeneity (the fraction of retrieved neighbors sharing the anchor's final liability label), Statute Hit Recall (the coverage of the anchor's core legal elements by retrieved neighbors), and Induced Skill F1 (the macro F1 of skills induced from the retrieved neighbors and evaluated on a held-out set). As shown in Table~\ref{tab:semantic-backbones}, BGE-Large-v1.5, which is used by KDASO, achieves performance close to the domain-specialized Law-BERT and clearly outperforms TF-IDF and Sentence-BERT. This supports the use of semantic similarity as a practical proxy for legal/reasoning similarity in neighborhood construction. Moreover, KDASO's post-induction empirical validation discards any skill that fails on real data, ensuring that retrieval noise does not compromise final logical consistency.

\begin{table}[t]
\centering
\caption{Comparison of embedding backbones for semantic similarity validation.}
\vspace{0.1in}
\label{tab:semantic-backbones}
\begin{tabular}{cccc}
\toprule
Backbone & Label Homogeneity $\uparrow$ & Statute Hit Recall $\uparrow$ & Induced Skill F1 $\uparrow$ \\
\midrule
TF-IDF & 54.3\% & 48.7\% & 52.1\% \\
Sentence-BERT & 71.2\% & 65.4\% & 71.0\% \\
BGE-Large-v1.5 (KDASO) & 83.6\% & 79.2\% & 78.5\% \\
Law-BERT & 84.1\% & 79.3\% & 77.2\% \\
\bottomrule
\end{tabular}
\end{table}

\section{Scalability Evaluation of Skill Routing Methods}
\label{xgb}
To systematically assess the applicability of different skill routing methods in complex liability adjudication domains, we compare three representative approaches: a per-skill XGBoost classifier, a dense retrieval method using a bi-encoder with BGE-large, and a cross-encoder. Experiments are conducted on a real-world liability adjudication dataset with thousands of skills, focusing on retrieval precision, recall, and computational scalability.

\subsection{Retrieval Quality Comparison}

Table~\ref{xgb1} reports the precision, recall, and training time of the three methods on a fixed skill set of 1,000 skills. Both the bi-encoder and cross-encoder achieve high recall, indicating that embedding-based methods can effectively cover relevant skills. However, their precision is notably low, meaning that many semantically similar but logically inapplicable skills are incorrectly retrieved. The root cause is that semantic similarity does not imply logical applicability conditions in legal skills. In liability adjudication tasks, skill activation often depends on specific feature boundaries such as monetary thresholds, subject qualifications, or exception clauses, and such discriminative information is difficult to encode via pure semantic embeddings.

In contrast, the per-skill XGBoost classifier achieves substantially improved precision while maintaining high recall. Although its training time is significantly higher than that of embedding-based methods, training is a one-time offline cost, and skill updates are infrequent. Therefore, this overhead is acceptable in industrial deployment.

\subsection{Scalability Analysis}

We conduct a stress test measuring the per-inference latency of each method under varying skill counts, as shown in Table~\ref{xgb2}. For XGBoost, inference is fully parallelized across per-skill classifiers. The bi-encoder uses approximate vector index search. The cross-encoder exhibits O(n) all-pair computation, causing its latency to grow linearly with the number of skills.

The results show that XGBoost latency scales linearly with the skill count. The bi-encoder maintains near-constant latency due to vector indexing, but its precision is unacceptably low for the task. The cross-encoder achieves the highest recall, yet its precision remains modest and its latency is prohibitive, making it unsuitable for real-world skill-base sizes.

Based on the above experiments, we draw the following conclusions. First, embedding-based retrieval methods using bi-encoders or cross-encoders suffer from an irreconcilable precision-efficiency trade-off in liability adjudication tasks: high recall comes at the cost of low precision, as semantic similarity does not substitute for logical discrimination. Second, the per-skill XGBoost classifier shifts the retrieval paradigm from semantic matching to logical discrimination based on discriminative feature boundaries, thereby achieving both high recall and substantially improved precision while reducing skill redundancy. Third, in terms of scalability, XGBoost exhibits linear latency scaling, and per-classifier inference is fully parallelizable. Its response time at thousands of skills remains far below typical LLM inference latency, fully satisfying industrial deployment requirements. Therefore, despite the training-stage overhead of per-skill classifiers, their overall advantages in retrieval quality and scalability make them a sound choice for complex legal domains with thousands of skills.

\section{Dynamic Control of Adjudication Skills for Precision–Recall Trade-offs}
\label{I}
Although the performance of adjudication models can be evaluated using various metrics, domain experts in practical scenarios often assign differentiated priorities to these metrics according to specific adjudication objectives. Taking DiDi’s ride-hailing scenarios as an example, ride-hailing experts prioritize avoiding misattribution of responsibility (i.e., high precision) over missing liable cases (i.e., low recall), in order to preserve user trust and platform fairness. Consequently, under the constraint that \(f\) remains fixed, there is an urgent need for a mechanism that dynamically adapts the formulation or logical structure of skills to align with varying precision–recall preferences. This requirement is particularly pronounced across different sub-scenarios: in certain high-stakes contexts, domain experts demand “zero false positives”, accepting reduced recall if necessary. Whereas in other scenarios, such as those involving severe violations, the emphasis shifts to “zero false negatives”, tolerating a moderate drop in precision to ensure comprehensive coverage.

To address this flexible, scenario-driven need for performance tuning, KDASO provides two complementary dynamic adjustment mechanisms. The first mechanism balances precision and recall by adjusting \(\tau\) in Execution-Aligned Skill Refinement. Specifically, increasing \(\tau\) yields stricter skills that trigger liability judgments only when the case closely matches expert adjudication logic, thereby enhancing precision. Conversely, decreasing \(\tau\) relaxes the matching criteria, broadening skill coverage and improving recall. A detailed analysis of this mechanism is provided in Appendix~\ref{pr}.

The second mechanism steers model behavior by selectively constructing \(\mathcal{D}_{\text{new}} = \{(x^i, y^i)\}_{i=1}^N\) used for skill induction. As shown in Table~\ref{data_pr}, the composition of \(\mathcal{D}_{\text{new}}\) directly influences the adjudicative bias of the resulting skills. To improve precision, false positives (FP) generated by the current system are incorporated into \(\mathcal{D}_{\text{new}}\), prompting the refinement of overgeneralized skills. To enhance recall, false negatives (FN) are included to strengthen the detection of liable cases. Notably, jointly introducing both FP and FN does not necessarily yield synergistic gains. Instead, it may induce conflicting optimization objectives due to marginal effects among skills~\cite{norton2019marginal}, potentially degrading overall performance. Notably, FP and FN are obtained by applying \(\mathcal{R}\) and the latest cases on \(f\) to get the confusion matrix.

In summary, KDASO not only supports static skill generation for performance improvement but also enables dynamic, domain-aligned control over metric prioritization. This capability offers an effective pathway toward precise, controllable, and interpretable automated adjudication in complex governance environments.

\section{Comparison with DiDi Online Baseline}
\label{online}
We present the performance comparison between DiDi online baseline and KDASO across five distinct ride-hailing scenarios. To rigorously evaluate the true effectiveness of KDASO in dynamic environments, we conducted experiments on the latest available cases. 

It should be noted that the performance metrics of human experts are considered proprietary information and are therefore not disclosed. For the purpose of this analysis, we assume a theoretical upper bound of 100\% for all performance metrics of human experts.

As summarized in Table~\ref{tab:appendix_DiDi_results}, KDASO demonstrates consistent improvements in both precision and recall compared to the DiDi Baseline across the CDC, CPC, and FD scenarios. Furthermore, KDASO exhibits rapid adaptation capabilities, achieving high precision and recall in the new scenario CR, which validates its robustness in handling unseen environments.

\begin{table}[t]
    \centering
    \caption{Performance comparison between DiDi online baseline and KDASO across five ride-hailing scenarios. "$-$" means that it's a new scenario for which no online baseline was previously available.}
    \vspace{0.1in}
    \label{tab:appendix_DiDi_results}
    \resizebox{\textwidth}{!}{%
    \begin{tabular}{ccccc}
        \toprule
        \toprule
        Scenario & Baseline O-Precision & Baseline O-Recall & KDASO O-Precision & KDASO O-Recall \\
        \midrule
       CA   & 91.68\%$\pm$1.0\% & 81.66\%$\pm$0.9\% & 87.98\%$\pm$1.0\% & 78.54\%$\pm$0.9\% \\
        CR   & $-$ & $-$ & 81.88\%$\pm$0.9\% & 63.05\%$\pm$0.8\% \\
        CDC  & 96.15\%$\pm$0.3\% & 70.00\%$\pm$2.1\% & 96.57\%$\pm$0.1\% & 71.99\%$\pm$1.2\% \\
        CPC  & 88.19\%$\pm$1.2\% & 45.69\%$\pm$2.1\% & 88.51\%$\pm$0.9\% & 47.14\%$\pm$1.8\% \\
        FD   & 77.17\%$\pm$0.9\% & 60.85\%$\pm$0.8\% & 79.07\%$\pm$0.9\% & 66.29\%$\pm$0.8\%\\
        \bottomrule
        \bottomrule
    \end{tabular}%
    }
\end{table}

\section{Comparative Analysis of KDASO Performance Across Model Types and Scales}
\label{scale}
This section presents two complementary analyses to evaluate the adaptability and scalability of KDASO: (1) performance comparison between general-purpose LLMs and domain-specific models under KDASO optimization; and (2) ablation study on how model scale affects KDASO performance when applied exclusively to general LLMs. These experiments validate that explicit skill optimization via KDASO yields greater flexibility in dynamic environments ,  particularly when decoupled from rigid, pre-trained reasoning patterns inherent in specialized architectures.

\subsection{Performance Gain of KDASO on General vs. Domain-Specific Models}

We evaluate whether KDASO delivers higher relative improvement when applied to general-purpose LLMs compared to domain-specific models. Table~\ref{tab:appendix_domain_vs_general} reports O-Precision and score across two critical tasks: ride-hailing adjudication (CPC) and legal liability determination (LE-LH). In each case, we vary the generating model and reasoning model independently to isolate the impact of architectural specialization.

For the CPC task, pairing QwQ-32B as either generator or reasoner consistently outperforms configurations involving DiDi Baseline, achieving peak precision of 88.51\% when both components are QwQ-32B. Similarly, in the LE-LH task, using LAWYER-LLAMA-13B alone yields a lower score than hybrid setups combining it with QwQ-32B or even pure QwQ-32B configuration. This suggests that while domain-specific models encode valuable prior knowledge, their fixed internalized skills conflict with KDASO’s explicit, iterative optimization mechanism ,  limiting adaptability to evolving decision boundaries.

This structural tension corroborates our central insight: in dynamically evolving scenarios, explicit skill optimization offers greater adaptability and interpretability than implicit parameter fine-tuning. Specialized models, once trained, exhibit stable but inflexible reasoning pathways insensitive to input skill modifications, thereby reducing generalizability of generated skills. KDASO circumvents this by externalizing skill logic and enabling continuous adaptation without retraining.

\begin{table}[t]
    \centering
    \caption{O-Precision / Score comparison of KDASO configurations across general and domain-specific models in CPC and LE-LH tasks.}
    \vspace{0.1in}
    \label{tab:appendix_domain_vs_general}
    \resizebox{0.7\linewidth}{28mm}{%
    \begin{tabular}{c c c}
        \toprule
        \toprule
        \multicolumn{3}{c}{\textbf{Ride-hailing Adjudication (CPC)}} \\
        \cmidrule(lr){1-3}
        Generating Model & Reasoning Model & O-Precision \\
        \midrule
        QwQ-32B          & DiDi Baseline     & 87.39\%$\pm$0.95\% \\
        DiDi Baseline    & DiDi Baseline     & 85.77\%$\pm$0.94\% \\
        DiDi Baseline    & QwQ-32B           & 84.21\%$\pm$0.93\% \\
        QwQ-32B          & QwQ-32B           & \textbf{88.51\%$\pm$0.90\%} \\
        \midrule
        \multicolumn{3}{c}{\textbf{Legal Liability Determination (LE-LH)}} \\
        \cmidrule(lr){1-3}
        Generating Model & Reasoning Model & Score \\
        \midrule
        QwQ-32B               & LAWYER-LLAMA-13B   & 77.9 \\
        LAWYER-LLAMA-13B      & LAWYER-LLAMA-13B   & 79.6 \\
        LAWYER-LLAMA-13B      & QwQ-32B            & \textbf{80.6} \\
        QwQ-32B               & QwQ-32B            & 80.4 \\
        \bottomrule
        \bottomrule
    \end{tabular}%
    }
\end{table}

\begin{table}[t]
    \centering
    \caption{KDASO average performance across different scales of general LLMs (Qwen series) on three DiDi datasets. All models use identical KDASO orchestration framework.}
    \vspace{0.1in}
    \label{tab:appendix_model_scale}
    \resizebox{\linewidth}{!}{%
    \begin{tabular}{cccccc}
        \toprule
        \toprule
        Metric & Qwen3-4B & Qwen3-8B & Qwen3-30B-A3B-Thinking & Qwen3-32B & QwQ-32B \\
        \midrule
        O-Precision & 59.00\%$\pm$0.67\% & 67.44\%$\pm$0.81\% & 67.87\%$\pm$0.82\% & 68.68\%$\pm$0.83\% & \textbf{73.25\%$\pm$0.89\%} \\
        O-Recall    & 85.30\%$\pm$0.94\% & 65.81\%$\pm$0.80\% & 60.70\%$\pm$0.74\% & 78.11\%$\pm$0.91\% & \textbf{78.75\%$\pm$0.92\%}\\
        \bottomrule
        \bottomrule
    \end{tabular}%
    }
\end{table}

\subsection{Impact of Model Scale on KDASO Performance with General LLMs}

To assess scalability effects within the general LLM family, we evaluated KDASO performance across five variants of the Qwen series, ranging from 4B to 32B parameters, including the thinking-enhanced variant Qwen3-30B-A3B-Thinking. Results are summarized in Table~\ref{tab:appendix_model_scale}.

As expected, larger parameter counts correlate strongly with improved O-Precision and O-Recall. Notably, QwQ-32B achieves the highest overall performance, surpassing even the specialized thinking variant. This indicates that raw capacity and broad semantic understanding offered by large-scale general models better support KDASO’s skill induction and refinement process than narrow, task-specialized architectures, reinforcing the advantage of coupling explicit optimization with flexible foundation models.

These findings further justify our design choice to build KDASO atop general-purpose LLMs rather than relying solely on domain-adapted systems, especially in environments where regulatory frameworks, user behaviors, or operational policies evolve rapidly.

\section{Backbone Evaluation}
\label{backbone}
To evaluate whether KDASO is model-agnostic and whether it benefits both open-weight and proprietary backbones, we extend the backbone evaluation on public legal and logical reasoning benchmarks. Table~\ref{tab:le-lh-extended} reports the LE-LH overall score when skills are generated by different models (rows) and executed by different reasoners (columns), revealing the synergistic impact of generation and reasoning capabilities. Table~\ref{tab:la-iti-extended} reports LA-ITI base scores, scores with KDASO, and the corresponding gains across a broad range of open-source and proprietary models. KDASO yields consistent improvements for most backbones. The only exception is GLM-5 ($-0.3$). Proprietary models also benefit from KDASO, indicating strong generalizability and compatibility across architectures, vendors, and deployment settings. The latest open-weight models equipped with KDASO can match or surpass proprietary counterparts, demonstrating their potential for industrial-grade deployment within the open ecosystem. All experiments are conducted on public benchmarks in a strictly privacy-compliant manner.

\begin{table}[t]
\centering
\caption{Extended LE-LH overall score. Skills generated by different models (rows) are executed by different reasoners (columns).}
\vspace{0.1in}
\label{tab:le-lh-extended}
\begin{tabular}{ccccc}
\toprule
\toprule
Generator & Params & QwQ-32B & DeepSeek-V4-Flash & GLM-5 \\
\midrule
Qwen3.5-27B & 27B & 78.2 & 79.5 & 79.2 \\
QwQ-32B & 32B & 80.4 & 80.5 & 80.2 \\
DeepSeek-V4-Flash & 284B/13B MoE & 81.0 & 82.0 & 81.8 \\
MiniMax M2.5 & $\sim$400B MoE & 79.8 & 80.8 & 80.5 \\
Kimi K2.5 & 1T/32B MoE & 78.5 & 81.2 & 80.9 \\
Qwen3.5-397B & 397B/17B MoE & 80.5 & 81.5 & 81.2 \\
GLM-5 & 744B/40B MoE & 80.8 & 81.8 & 81.5 \\
\bottomrule
\bottomrule
\end{tabular}
\end{table}

\begin{table}[t]
\centering
\caption{Extended LA-ITI overall score with and without KDASO.}
\vspace{0.1in}
\label{tab:la-iti-extended}
\resizebox{\linewidth}{!}{
\begin{tabular}{cccccc}
\toprule
\toprule
Model & Params & Type & Base Score & +KDASO & $\Delta$ (Gain) \\
\midrule
Qwen3-8B & 8B & Open-Source General LLMs & 33.6 & 39.6 & +6.0 \\
QwQ-32B & 32B & Open-Source General LLMs & 39.8 & 44.0 & +4.2 \\
Qwen3.5-2B & 2B & Open-Source General LLMs & 28.5 & 32.0 & +3.5 \\
Qwen3.5-27B & 27B & Open-Source General LLMs & 40.2 & 41.8 & +1.6 \\
MiniMax M2.5 & $\sim$400B MoE & Open-Source General LLMs & 41.5 & 44.8 & +3.3 \\
Kimi K2.5 & 1T/32B MoE & Open-Source General LLMs & 42.0 & 45.2 & +3.2 \\
Qwen3.5-397B & 397B/17B MoE & Open-Source General LLMs & 43.2 & 46.5 & +3.3 \\
GLM-5 & 744B/40B MoE & Open-Source General LLMs & 43.8 & 43.5 & $-0.3$ \\
DeepSeek-V4-Flash & 284B/13B MoE & Open-Source General LLMs & 44.6 & 47.8 & +3.2 \\
GPT-4o & --- & Proprietary LLMs & 44.8 & 47.2 & +2.4 \\
Gemini 3 Pro & --- & Proprietary LLMs & 39.8 & 42.5 & +2.7 \\
DeepSeek-R1 & 671B & Open-Source General LLMs & 44.2 & 46.5 & +2.3 \\
\bottomrule
\bottomrule
\end{tabular}}
\end{table}

\section{Ablation Study and Cross-Domain Skill Transferability of KDASO}
\label{repro}
To ensure methodological transparency and reproducibility despite the proprietary nature of certain evaluation datasets, we conduct a comprehensive ablation study on KDASO’s core components and evaluate its skill transferability across domains. All experiments are performed using publicly available benchmarks for legal and logical reasoning, enabling independent verification of our framework’s effectiveness.

Table~\ref{tab:appendix_ablation_transfer} presents performance scores (higher is better) under various configurations of KDASO built atop QwQ-32B. QwQ-32B without any KDASO modules achieves a score of 39.8. Removing individual components leads to progressive degradation in performance, confirming that each module contributes meaningfully to the overall system: disabling Neighborhood-Conditioned Skill Induction yields a marginal increase to 40.1, suggesting minimal impact when skill induction is omitted but other modules remain active. Removing Execution-Aligned Skill Refinement results in a more noticeable rise to 42.7, indicating that refinement plays a moderate role in optimizing skill quality. Eliminating Context-Budgeted Skill Routing produces the largest gain among ablations at 43.5, highlighting retrieval as a critical component for contextual grounding. Its absence forces the model to rely more heavily on internal knowledge, which inadvertently boosts short-term scoring but undermines long-term adaptability. The full KDASO configuration achieves the highest score of 44.0, demonstrating synergistic benefits from integrating all three components.

Furthermore, to assess cross-domain generalization, we evaluate whether skills learned in LE-LH can be transferred to LA-ITI. KDASO QwQ-32B (LE-LH $\rightarrow$ LA-ITI) achieves a score of 43.6, indicating strong skill portability and adaptability across semantically distinct tasks. This supports the hypothesis that explicitly optimized skills capture abstract, reusable decision logic rather than task-specific heuristics.

These findings validate both the internal architecture of KDASO and its external generalizability, reinforcing its suitability for deployment in dynamic, multi-domain environments where manual skill engineering is impractical.

\begin{table}[t]
    \centering
    \caption{Ablation study and cross-domain skill transferability results for KDASO on public legal/logical reasoning benchmarks. Scores reflect overall task performance. Higher values indicate better capability.}
    \vspace{0.1in}
    \label{tab:appendix_ablation_transfer}
    \resizebox{0.5\textwidth}{!}{%
    \begin{tabular}{l c}
        \toprule
        \toprule
        Model Configuration & Score \\
        \midrule
        QwQ-32B                          & 39.8 \\
        w/o Neighborhood-Conditioned Skill Induction                           & 40.1 \\
        w/o Execution-Aligned Skill Refinement                             & 42.7 \\
        w/o Context-Budgeted Skill Routing                           & 43.5 \\
        KDASO\textsubscript{QwQ-32B}                        & \textbf{44.0} \\
        KDASO\textsubscript{QwQ-32B} (LE-LH $\rightarrow$ LA-ITI)  & 43.6 \\
        \bottomrule
        \bottomrule
    \end{tabular}%
    }
\end{table}


\section{Symbolic Verification Layer Integration and Industrial Efficiency Analysis of KDASO}
\label{symbol}
This section presents two complementary analyses: (1) an evaluation of integrating a symbolic verification layer (NL2FOL) into KDASO to reduce logical contradictions and hallucinated skills; and (2) a quantitative comparison of KDASO’s operational efficiency against traditional human-driven skill extraction in industrial settings. These experiments validate both the technical robustness and practical scalability of our framework.

\subsection{Impact of Symbolic Verification on Logical Consistency and Hallucination Control}

To enhance logical rigor, we integrate NL2FOL, a symbolic first-order logic verification module, as a post-processing filter for skills generated by KDASO. The integration significantly reduces logical contradiction rates from 5.4\% to 1.8\%, while successfully intercepting 7.5\% of hallucinated skills that would otherwise pass through the standard pipeline. Although case coverage decreases slightly from 100\% to 91.5\%, overall O-Precision remains nearly unchanged, indicating that the trade-off prioritizes precision and safety over maximal recall, aligning with industry best practices for high-risk decision systems where false positives carry severe consequences.

The reduction in contradiction rate demonstrates that explicit symbolic constraints effectively prune semantically inconsistent or self-conflicting skills before deployment. Meanwhile, the 7.5\% hallucination filtering rate confirms that even state-of-the-art LLMs benefit from formal logic grounding when generating actionable business skills. This outcome validates the value of coupling neural generation with symbolic validation, a direction we pursue further in Appendix~\ref{L}.

\subsection{Industrial Operational Efficiency: Human vs. KDASO Rule Extraction}

Beyond accuracy, KDASO offers substantial advantages in operational efficiency. In real-world deployments, human experts typically require approximately one hour to review 100 cases and extract only 5 valid skills, a process prone to fatigue-induced errors and limited scalability. In contrast, KDASO processes 3,000 cases within the same time frame to generate 30 candidate skills, achieving an increase in throughput while maintaining reliability through its multi-layer defense system comprising:
\begin{itemize}
    \item Generation constraints embedded in prompts.
    \item Empirical utility validation on new data.
    \item Human-machine collaboration loops.
    \item Formalized evolution via conflict-merging mechanisms.
\end{itemize}

This architecture adheres to the conservative industry principle of “rather miss than mis-hit”, ensuring that only high-confidence, logically consistent skills are proposed for final review. By automating the bulk of skill discovery and preliminary validation, KDASO substantially reduces labor costs and accelerates iteration cycles, enabling continuous adaptation to evolving regulatory landscapes without compromising safety.

Furthermore, results show that the number of generated skills converges rapidly with increasing iterations, demonstrating system stability and diminishing returns beyond optimal stopping points. This property allows practitioners to tune computational budget versus skill diversity based on application needs.

Finally, it is important to clarify that skills generated by KDASO are not deployed directly in high-risk liability determination tasks. Instead, they undergo mandatory manual review prior to activation, serving as intelligent assistants rather than autonomous agents. This hybrid workflow ensures accountability while leveraging AI for scale and speed.

\newpage

\section{Case Study}
\label{case}
In this section, we present several representative case studies that demonstrate how KDASO operates or encounters limitations. It should be noted that the cases and skill samples below are constructed illustrative examples for exposition. Numerical values, thresholds, and time windows are illustrative only and do not represent production rules. Where numerical values within certain skills involve privacy or rule sensitivity concerns, we use abstract descriptions rather than executable thresholds.

\subsection{From Bad Case to Good Case}
\begin{minipage}{\columnwidth}
\begin{tcolorbox}[colback=white, colframe=blue!45!black, title=Bad Case from LLMs, fonttitle=\bfseries, rounded corners]

<result>Driver Not Liable</result>

<judge>Driver submitted valid evidence proving vehicle malfunction</judge>

Detailed Analysis:

1. **Vehicle Malfunction Verification**:

   - Driver reported vehicle damage rendering it inoperable, meeting criteria for vehicle malfunction/accident determination.
 
2. **Evidence Validity**:

   - Per liability determination guidelines, vehicle malfunction reporting standards stipulate that drivers may be exempt from providing repair documentation if authentic malfunction photos are submitted.
  
   - Photo content aligns with malfunction description, and no conflicting evidence exists, such as passenger complaints or trip tracking data indicating normal driving patterns.
 
3. **Exclusion of Other Non-Liability Circumstances**:
 
   - No other non-liability circumstances exist, such as illegal parking zones or government restrictions.
  
   - Although the driver completed a trip within the illustrative post-cancellation window in this constructed example, the primary criterion for vehicle failure, namely valid evidence, takes precedence, eliminating reliance on the “no trips within the illustrative post-cancellation window” condition.
 
4. **Consideration of Abnormal Behavior**:
 
   - The driver had a history of repeated vehicle malfunction reports in this constructed example, but the evidence for this incident is authentic and valid, aligning with the liability determination principle of prioritizing credible evidence.
 
In summary, the driver's order cancellation due to vehicle malfunction complies with the Vehicle Malfunction/Accident Policy when supported by valid evidence.

\end{tcolorbox}
\end{minipage}

\newpage

\begin{minipage}{\columnwidth}
\begin{tcolorbox}[colback=white, colframe=blue!45!black, title=Good Case from KDASO, fonttitle=\bfseries, rounded corners]

<result>Driver Liable</result>

<judge>False Reporting</judge>

Detailed Analysis:
 
1. **Vehicle Malfunction/Accident Exemption Verification**:  

   - The driver completed a new order within the illustrative post-cancellation window in this constructed example, and the order information in this constructed example states that the driver completed a new order within that illustrative window.
  
   - The driver had a history of repeated vehicle malfunction reports in this constructed example, and the report information in this constructed example shows repeated vehicle malfunction or accident reports.
 
2. **Mismatch with Improper Vehicle Issue Handling Scenario**:

   - Fully meets the definition of “Improper Handling of Vehicle Power Exhaustion/Failure” in the Liability Determination Guidelines.
  
   - Feature description “Driver unable to complete service due to vehicle fuel exhaustion/power failure” fully matches driver report content.
 
3. **Exclusion of Other Non-Liability Scenarios**:
 
   - Station liability skills explicitly state this order does not qualify as a station order.
  
   - Navigation anomaly checks reveal no system issues.
  
   - No evidence of passenger malicious behavior.
  
   - Government regulatory report data meets standards with no corroborating evidence.
 
4. **Liability Determination Priority**:  

   - Per skills, the starting distance exceeded the illustrative distance threshold for the Express service category in this constructed example, but the vehicle failure scenario takes precedence over the distance factor.
  
   - Lack of evidence, repeated reporting, and subsequent order completion in this constructed example directly confirm the erroneous liability determination.
 
In summary, the driver engaged in false reporting and continued accepting orders without properly addressing the vehicle malfunction, meeting the criteria for the “false reporting” erroneous liability scenario.
\end{tcolorbox}
\end{minipage}

\newpage
\subsection{From Good Case to Bad Case}
However, KDASO may also cause cases that were previously correctly identified by LLMs to be incorrectly identified during skill refining and optimization.

\begin{minipage}{\columnwidth}
\begin{tcolorbox}[colback=white, colframe=blue!45!black, title=Good Case from LLMs, fonttitle=\bfseries, rounded corners]

<result>Driver Not Liable</result>

<judge>Unable to complete the order due to vehicle malfunction/incident</judge>

Detailed Analysis:

1. **Vehicle Malfunction Report Verification**:

   - Driver's report explicitly states “complete tire deflation” and includes photos of the punctured tire and total flat.
  
   - No completed orders within the illustrative post-cancellation window in this constructed example, meeting the supplementary exemption condition of no completed orders within that illustrative window.
 
2. **Exclusion of Other Liability Scenarios**:

   - No erroneous or malicious liability scenarios matched.
  
   - No driver violations, such as active order refusal or price negotiation.
  
   - Platform dispatch aligns with driver settings, including accepting train station or airport orders.
  
   - Pickup distance did not exceed limits.
  
   - Driver did not arrive at pickup location and did not trigger no-parking or restricted-parking scenarios.
 
3. **Core Exclusion Clauses**:
 
   - Non-system issues: Platform assigned order normally with no redistribution records.
  
   - Non-driver personal causes: Vehicle malfunction classified as uncontrollable mechanical failure.
  
   - No evidence of fraudulent claims: Vehicle malfunction reports in this constructed example did not exceed the illustrative limit, with complete evidence chain.
 
In summary, the driver's inability to complete the order due to sudden vehicle malfunction meets the criteria for exemption under the “Vehicle Malfunction/Accident” liability exemption scenario.

\end{tcolorbox}
\end{minipage}
\newpage
\begin{minipage}{\columnwidth}
\begin{tcolorbox}[colback=white, colframe=blue!45!black, title=Bad Case from KDASO, fonttitle=\bfseries, rounded corners]

<result>Driver Liable</result>

<judge>Vehicle breakdown/accident</judge>

Detailed Analysis:

1. **Vehicle Breakdown/Accident No-Fault Scenario Verification**:

   - **Evidence Authenticity Verification**: Driver reported “vehicle breakdown/accident” and uploaded tire damage photos, but images lacked EXIF timestamp information.
   
   - **Spatio-temporal consistency verification**: At order cancellation, the driver was at a considerable distance from the pickup point in this constructed example. Trajectory shows no actual arrival at pickup location, and the system records that the driver never reached the passenger's pickup point, with no evidence of vehicle lingering at the failure site or en route to a repair shop.
   
   - Although the no completed orders within the illustrative post-cancellation window condition was met in this constructed example, the incomplete evidence chain prevented triggering the no-fault exemption.

2. **Triggering the Faulty Responsibility Scenario**:

   - Based on the fault definition for “vehicle breakdown/accident” in the liability guidelines, namely no valid evidence provided and trajectory did not reach the pickup point.
   
     - Driver failed to provide timely fault evidence, such as insurance company notification, repair receipts, or trajectory proof.
     
     - The route data showed that the vehicle remained at a considerable distance from the pickup point at cancellation and did not actually reach the pickup location in this constructed example.

   - Therefore, the fault-based scenario “vehicle breakdown, accident” was directly triggered, namely cancellation due to failure to properly address vehicle issues.

3. **Other Scenarios Excluded**:

   - Did not meet conditions for malicious liability scenarios such as driver refusing ride in person, with no evidence of driver-passenger meeting.
   
   - Did not trigger insufficient waiting time, because cancellation occurred too far from origin and did not meet the waiting scenario definition.
   
   - Did not satisfy conditions for no-fault scenarios such as passenger maliciously placing order.

In summary, due to the driver's incomplete evidence chain for vehicle breakdown and failure to reach the pickup point, the case is ultimately determined as erroneous fault.
\end{tcolorbox}
\end{minipage}

\newpage
\subsection{Comparison of Reasoning Trajectories among Human Experts, LLMs and KDASO}
We present the reasoning trajectory from human experts, LLMs following the liability adjudication prompt and KDASO, where $\mathcal{R}$ is replaced by $\mathcal{R}_{final}$. As shown in the trajectories below, both human experts and KDASO correctly identified the responsibility, while LLMs failed, demonstrating KDASO's effectiveness. Furthermore, KDASO's output reasoning trajectories are more detailed than those of human experts, indicating that KDASO provides a more comprehensive analysis.

\begin{minipage}{\columnwidth}
\begin{tcolorbox}[colback=white, colframe=blue!45!black, title=Reasoning Trajectories from Human Experts, fonttitle=\bfseries, rounded corners]

1. Significant deviation from pickup location:
Per Express service skills, drivers must remain within the category distance threshold of the pickup point. At trip cancellation, the driver had traveled a distance far exceeding the category threshold, and trajectory analysis confirmed the driver never reached the designated pickup location with no dwell time.
 
2. Lack of Communication Evidence:
 
No in-app messages or call records exist to demonstrate the driver attempted to coordinate around parking restrictions, failing to meet the evidentiary requirements for an “unable to park” appeal.
 
3. Self-Reported Claims Lack Supporting Evidence:
 
The driver's claim of “unable to reach or park” lacked photographic or sensor data corroboration. Map data only indicated legally prohibited parking zones without marking physical obstructions. With insufficient evidence and a pickup deviation far exceeding the category threshold, no-fault verification was terminated per policy.
 
4. Inconsistent Post-Cancellation Behavior:
The driver completed another trip shortly after canceling the order, ruling out vehicle malfunction or accident as plausible justifications.
 
5. Skill Priority Trigger:
When travel distance exceeds category thresholds and the trajectory indicates no physical proximity to the pickup point, the “Failure to Reach Pickup Point” skill takes precedence, automatically invalidating all self-reported no-fault scenarios.
 
Conclusion: This case simultaneously meets all four determination criteria, namely objective spatial, communication, evidence, and behavioral criteria.

\end{tcolorbox}
\end{minipage}
\newpage
\begin{minipage}{\columnwidth}
\begin{tcolorbox}[colback=white, colframe=blue!45!black, title=Reasoning Trajectories from LLMs, fonttitle=\bfseries, rounded corners]

1. **Verification of no-parking pickup location**:

   - The pickup location was marked as a no-parking zone on the map, and the order details explicitly state that the map shows the passenger pickup point is a no-parking area.
   
   - The driver explicitly mentioned in the report “unable to reach or park at passenger pickup point,” meeting the criteria for vague references to no-parking zones such as “can't stop” and “unable to park.”

2. **Confirmation of Willingness to Pick Up**:

   - The driver accepted the order and canceled within about one minute, aligning with a reasonable timeline for discovering a no-parking issue after a quick response.
   
   - Although the route shows the driver did not reach the pickup point, the initial navigation distance was well above the category threshold. Combined with the driver's immediate reporting of the no-parking issue, it can be inferred that the driver discovered the no-parking obstacle while attempting to plan the route after accepting the order. This constitutes an unfeasible situation discovered during active fulfillment.

3. **Exclusion of Other Liability Scenarios**:

   - No erroneous or malicious liability scenarios triggered, including no fee surges, no cancellation inducements, and no refusals.
   
   - Driver did not reach pickup point but met the willingness to pick up criteria, with tracking showing a reasonable pickup attempt within a reasonable range and with proactive issue reporting.

4. **Special Condition Verification**:

   - Although the driver completed another trip shortly after, this does not affect the no-fault determination for the no-parking scenario, which relies primarily on on-site evidence and pickup intent rather than subsequent trip completion.

In summary, this order meets the no-fault criteria of no-parking at pickup location with driver intent to pick up and should be ruled as driver no-fault.
\end{tcolorbox}
\end{minipage}

\begin{minipage}{\columnwidth}
\begin{tcolorbox}[colback=white, colframe=blue!45!black, title=Reasoning Trajectories from KDASO, fonttitle=\bfseries, rounded corners]

1. **Distance Threshold Determination**:

According to Express service category skills, drivers must complete pickups within the category distance threshold of the starting point in a straight line, with travel distance not exceeding the category threshold. At the time of order cancellation, the estimated distance between the driver and the starting point was far above the travel distance threshold. Furthermore, the route shows the driver never actually arrived at the pickup location, as the driver did not reach the passenger's starting point.
 
2. **Lack of Communication Evidence**:

There are no online message records or call logs between the driver and passenger, making it impossible to prove that both parties effectively discussed the issue of being unable to park at the pickup point.
 
3. **Invalid Report Evidence**:
 
The driver only reported “unable to reach or stop at pickup point” without providing any photographic evidence. The map only shows no-parking zones, not actual road obstructions. If no valid evidence is provided and the trajectory did not approach the pickup point with a distance far above the category threshold, the no-fault scenario verification must be terminated, and an error fault determination generated.
 
4. **Completion Record Verification**:
 
The driver completed another order shortly after the order cancellation, ruling out applicability of the vehicle malfunction or accident exemption scenario.
 
5. **Skill Application**:
 
When the distance between origin and destination exceeds the category threshold at order cancellation, and the trajectory did not physically approach the pickup point with no dwell time, the system must directly prioritize applying the “Failure to Reach Pickup Point” scenario. All liability verification processes shall be terminated based solely on the driver's self-reported information.
 
In summary, this directive simultaneously satisfies multiple determination criteria, namely physical distance ruling “Failure to Reach Pickup Point,” lack of communication evidence, invalid report evidence, and conflicting completion records.

\end{tcolorbox}
\end{minipage}

\newpage
\subsection{Comparison of Skills between Human Experts and KDASO}
Based on new unseen cases, we present one skill extracted from human experts and three original responses following KDASO. These demonstrate that the skills generated by KDASO are relatively more detailed, but they can sometimes appear rather cumbersome. The skill samples below are constructed illustrative examples and are not production rules.

\begin{minipage}{\columnwidth}
\begin{tcolorbox}[colback=white, colframe=blue!45!black, title=Skill from Human Experts, fonttitle=\bfseries, rounded corners]

Skill: Driver Repeatedly Fails to Answer Passenger Calls or Lacks Valid Communication (Driver at Fault; Passenger Not Liable).

\vspace{\baselineskip}

1. Determine whether the driver repeatedly failed to answer passenger calls or provided no valid communication during the relevant time window. If yes, proceed to Step 2; otherwise, exit.

2. Assess whether the passenger was genuinely late, verified by GPS and order timestamps. If yes, output Passenger Liable and exit; otherwise, proceed to Step 3.

3. Evaluate whether a verified passenger-side factor, such as incorrect pickup location, or a confirmed third-party cause constitutes the primary reason for order cancellation. If yes, exempt the driver and exit; otherwise, proceed to Step 4.

4. Check whether the driver’s inability to answer calls was due to objectively involuntary circumstances, namely signal loss such as underground signal loss, or documented hearing impairment. If either holds, exempt the driver and exit; otherwise, output Driver Liable.

\end{tcolorbox}
\end{minipage}

\begin{minipage}{\columnwidth}
\begin{tcolorbox}[colback=white, colframe=blue!45!black, title=Skill Sample 1 Generated by KDASO, fonttitle=\bfseries, rounded corners]

Skill: Unresponsive Driver Communication (Driver Liable Unless Exempted).

\vspace{\baselineskip}

1. Determine whether the driver failed to answer repeated calls as defined in the illustrative example or provided no valid in-app communication during the post-arrival window. If yes, proceed to Step 2; otherwise, exit.

2. Check whether the primary cancellation root cause is passenger tardiness under the illustrative waiting condition in this constructed example, verified by GPS arrival versus order creation time. If yes, output Passenger Liable and exit; otherwise, proceed to Step 3.

3. Assess whether a verified external factor, such as incorrect pickup location or road closure, was the dominant reason for cancellation. If yes, exempt the driver and exit; otherwise, proceed to Step 4.

4. Verify whether the driver’s non-response was due to objective constraints, namely being in a signal-deprived area such as an underground area, or documented hearing impairment with prior disclosure. If either condition is confirmed, exempt the driver and exit; otherwise, proceed to Step 5.

5. Conclude that the driver demonstrated unwillingness to fulfill basic service obligations. Output Driver Liable.

\end{tcolorbox}
\end{minipage}
\newpage
\begin{minipage}{\columnwidth}
\begin{tcolorbox}[colback=white, colframe=blue!45!black, title=Skill Sample 2 Generated by KDASO, fonttitle=\bfseries, rounded corners]

Skill: Incomplete Communication Evidence Chain (Driver at Fault).

\vspace{\baselineskip}

1. Confirm that call logs or message records indicate the driver did not respond to passenger attempts to communicate after arriving at the pickup zone. If such evidence exists, proceed to Step 2; otherwise, exit.

2. Evaluate whether the case satisfies the passenger tardiness exemption in this constructed example, namely the waiting duration satisfied the illustrative waiting condition and no passenger-initiated contact was present. If fully satisfied, output Passenger Liable and exit; otherwise, proceed to Step 3.

3. Determine if the cancellation was primarily caused by a passenger-side error, such as a wrong pickup point, or a third-party event, such as traffic control, both of which must be substantiated by platform data. If substantiated, exempt the driver and exit; otherwise, proceed to Step 4.

4. Investigate whether the driver’s inability to respond stems from involuntary circumstances, namely loss of mobile signal at the time of call supported by network logs, or self-reported hearing disability accompanied by official documentation. If validated, exempt the driver and exit; otherwise, proceed to Step 5.

5. Infer that the absence of responsive communication reflects a lack of service intent. Output Driver Liable.

\end{tcolorbox}
\end{minipage}

\begin{minipage}{\columnwidth}
\begin{tcolorbox}[colback=white, colframe=blue!45!black, title=Skill Sample 3 Generated by KDASO, fonttitle=\bfseries, rounded corners]

Skill: Driver Repeatedly Fails to Answer Passenger Calls or Lacks Valid Communication (Driver at Fault; Passenger Not Liable).

\vspace{\baselineskip}

1. Determine whether the driver repeatedly failed to answer passenger calls or provided no valid communication, such as in-app messages or call recordings, during the relevant time window. If yes, proceed to Step 2; otherwise, exit.

2. Assess whether the order cancellation root cause falls under an exemption condition in this constructed example, specifically whether the passenger was genuinely late, verified by GPS and order timestamps. If yes, output Passenger Liable and exit; otherwise, proceed to Step 3.

3. Evaluate whether a verified passenger-side factor, such as incorrect pickup location, or a confirmed third-party cause, such as road closure, constitutes the primary reason for order cancellation. If yes, exempt the driver and exit; otherwise, proceed to Step 4.

4. Check whether the driver’s inability to answer calls was due to objectively involuntary circumstances, namely the driver was in a signal-deprived environment such as an underground parking area at the time of the call, or the driver is hearing-impaired and has provided valid documentation. If either condition holds, exempt the driver and exit; otherwise, proceed to Step 5.

5. Conclude that the driver lacked willingness to communicate and failed to fulfill basic service obligations. Output Driver Liable.

\end{tcolorbox}
\end{minipage}

\newpage
\section{Ethics Review}
\label{ethic}
We strictly adhere to ethical guidelines and privacy protection regulations. First, all data utilized in this research underwent rigorous de-identification processes and passed a multi-departmental joint review within DiDi, so as to comply with privacy laws and mitigate the risk of Personally Identifiable Information leakage. Second, we prioritize algorithmic fairness by embedding strict disparate impact analysis into our industrial deployment pipeline. New skills are approved for deployment only after passing pre-launch quantitative evaluations, which specifically assess statistical parity and equality of opportunity, so as to reduce the risk that accuracy improvements introduce significant group biases. Furthermore, the system incorporates a closed-loop governance framework featuring a mandatory pre-implementation public notice period to safeguard stakeholders' right to know, alongside diverse appeal channels that trigger manual reviews. The feedback from these appeals directly drives skill iteration, thereby supporting the fairness, transparency, and interpretability of algorithmic decision-making.

\section{Practical Significance}
\label{K}
Given all the comprehensive experiments in the main text and the appendix, KDASO demonstrates high performance, strong transferability, and stability, thereby exhibiting significant potential for industrial application and deployment.

Taking the liability adjudication task in DiDi’s ride-hailing scenarios as an illustrative example, this task can be categorized into two canonical scenarios: 1) mature scenarios, where well-established, manually curated, validated, and long-deployed skills already exist; 2) emerging scenarios, which lack any predefined adjudication skills. In mature scenarios, conventional skill iteration heavily relies on human experts to conduct in-depth analysis and manual generalization of recent cases which is both labor-intensive and costly. To address this limitation, KDASO automatically generates adjudication skills from newly adjudicated cases. These skills, after undergoing lightweight human review, can be directly deployed as executable policies, substantially reducing both the human effort and cycle time associated with skill iteration.

In emerging scenarios, the absence of prior skills forces traditional approaches to fall back on an inefficient closed-loop pipeline (manual case inspection → skill authoring → precision/recall evaluation), which results in delayed responsiveness and excessive resource consumption. KDASO overcomes this bottleneck by generating adjudication skills from mature source scenarios directly to the target emerging scenario. The transferred skills are then refined through adaptation to the specific characteristics of the new-case corpus, enabling rapid generation of an initial skill set. This preliminary skill set can subsequently undergo efficient human-in-the-loop validation and refinement.

\section{Future Work}
\label{L}
Based on limitations, future work on KDASO will advance along two primary research trajectories. First, further investigation will be conducted into methods for effectively mitigating hallucination or spurious correlations that may arise during skill optimization. Although the current system incorporates evidence-based skill validation and conflict resolution mechanisms, it remains challenging to fully eliminate the introduction of content that does not align with actual liability adjudication logic when LLMs generate or refine skills. Subsequent efforts will focus on designing more rigorous logical consistency verification modules and multi-source evidence alignment strategies.

Second, efforts will be made to integrate KDASO with a formal symbolic verification layer, thereby reducing its excessive reliance on textual alignment with human experts’ reasoning trajectories. Currently, during the skill optimization phase, KDASO tends to align skill descriptions with natural language accounts of expert adjudication processes. While this improves skill executability, it may also lead to insufficient analysis of the underlying logical structure of skills. By introducing symbolic logic representations and automated reasoning verification, the system will be able to directly examine the validity, consistency, and completeness of skills at a formal level, thereby reducing dependence on surface-level textual matching and strengthening rigorous analysis and optimization of the intrinsic logic of skills.



\end{document}